\documentclass[10pt,twocolumn,letterpaper]{article}

\usepackage[pagenumbers]{cvpr} 

\usepackage{graphicx}
\usepackage{amsmath}
\usepackage{amssymb}
\usepackage{booktabs}
\usepackage[utf8]{inputenc} 
\usepackage[T1]{fontenc}    
\usepackage{url}            
\usepackage{booktabs}       
\usepackage{amsfonts}       
\usepackage{nicefrac}       
\usepackage{microtype}      
\usepackage{xcolor}         
\usepackage[linesnumbered,ruled,vlined]{algorithm2e}
\usepackage{array}
\usepackage{amssymb}
\usepackage{latexsym}
\usepackage{amsmath,amssymb}
\usepackage{graphicx}
\usepackage{multirow}
\usepackage{adjustbox}
\usepackage{makecell}
\usepackage[nottoc]{tocbibind}\usepackage{url}
\usepackage{wrapfig}

\SetCommentSty{mycommfont}
\newlength{\commentWidth}
\usepackage[pagebackref,breaklinks,colorlinks]{hyperref}

\usepackage{amsmath,amsfonts,bm}

\def\eqref#1{equation~\ref{#1}}

\def\1{\bm{1}}

\DeclareMathAlphabet{\mathsfit}{\encodingdefault}{\sfdefault}{m}{sl}
\SetMathAlphabet{\mathsfit}{bold}{\encodingdefault}{\sfdefault}{bx}{n}

\newcommand{\y}{\boldsymbol{y}}

\newcommand{\cb}{{\boldsymbol c}}

\newcommand{\nb}{{\boldsymbol n}}

\newcommand{\qb}{{\boldsymbol q}}

\newcommand{\xb}{{\boldsymbol x}}

\newcommand{\zb}{{\boldsymbol z}}

\newcommand{\beq}{\begin{equation}}
\newcommand{\eeq}{\end{equation}}
\newcommand{\beqa}{\begin{eqnarray}}
\newcommand{\eeqa}{\end{eqnarray}}

\newcommand{\epsilonb}{\boldsymbol{\epsilon}}

\newcommand{\textred}[1] {\textcolor{red}{#1}} 

\usepackage[capitalize]{cleveref}
\crefname{section}{Sec.}{Secs.}
\Crefname{section}{Section}{Sections}
\Crefname{table}{Table}{Tables}
\crefname{table}{Tab.}{Tabs.}

\def\cvprPaperID{6435} 
\def\confName{CVPR}
\def\confYear{2023}

\begin{document}

\title{DATID-3D: Diversity-Preserved Domain Adaptation \\ Using Text-to-Image Diffusion for 3D Generative Model}

\author{Gwanghyun Kim$^1$ \qquad \stepcounter{footnote}Se Young Chun$^{1,2,}$\thanks{} \\
$^1$Dept. of Electrical and Computer Engineering, $^2$INMC \& IPAI \\
Seoul National University, Republic of Korea\\
{\tt\small \{gwang.kim, sychun\}@snu.ac.kr}
}

\twocolumn[{%
\maketitle
\renewcommand\twocolumn[1][]{#1}%
\begin{center}
    \centering
    \captionsetup{type=figure}
    \vspace{-2em}
    \includegraphics[width=1\textwidth]{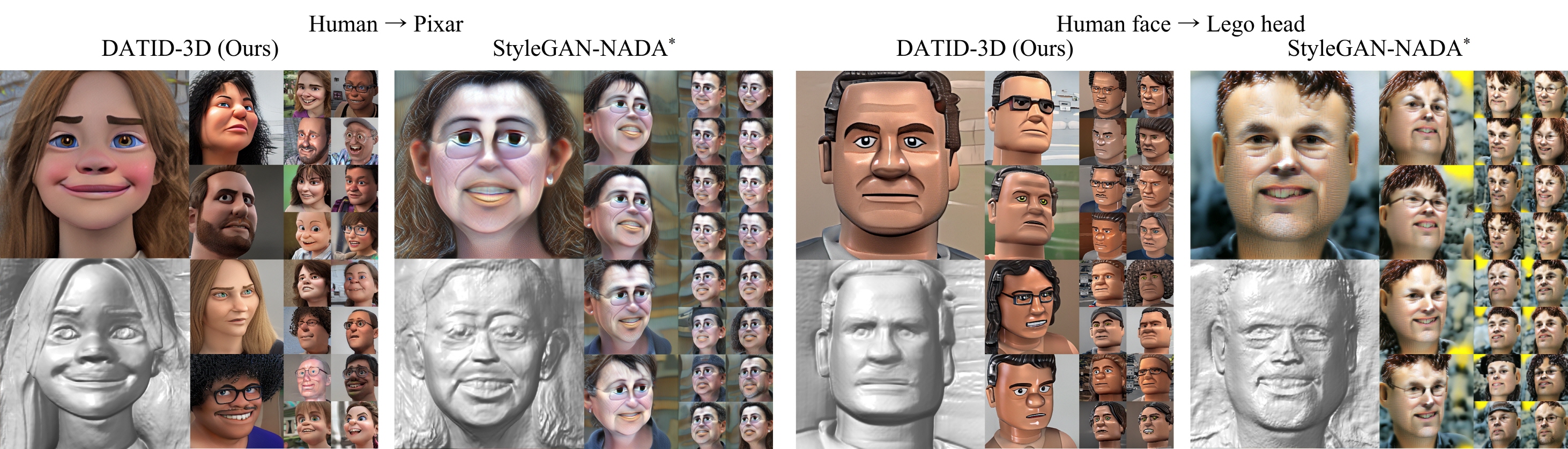}
    \vspace{-1.5em}
    \captionof{figure}{Our DATID-3D succeeded in domain adaptation of 3D-aware generative models without additional data for the target domain while preserving diversity that is inherent in the text prompt as well as enabling high-quality pose-controlled image synthesis with excellent text-image correspondence.
    However, StyleGAN-NADA$^*$, a 3D extension of the state-of-the-art StyleGAN-NADA for 2D generative models~\cite{gal2021stylegan}, yielded alike images in style with poor text-image correspondence.
    See the supplementary videos at \href{https://gwang-kim.github.io/datid_3d}{gwang-kim.github.io/datid\_3d}. 
    }
    \label{fig1_overview}
\end{center}%
}]
{
  \renewcommand{\thefootnote}%
    {\fnsymbol{footnote}}
  \footnotetext[1]{Corresponding author.}
}


\begin{abstract}
Recent 3D generative models have achieved remarkable performance in synthesizing high resolution photorealistic images with view consistency and detailed 3D shapes, but training them for diverse domains is challenging since it requires massive training images and their camera distribution information. Text-guided domain adaptation methods have shown impressive performance on converting the 2D generative model on one domain into the models on other domains with different styles by leveraging the CLIP (Contrastive Language-Image Pre-training), rather than collecting massive datasets for those domains. However, one drawback of them is that the sample diversity in the original generative model is not well-preserved in the domain-adapted generative models due to the deterministic nature of the CLIP text encoder. Text-guided domain adaptation will be even more challenging for 3D generative models not only because of catastrophic diversity loss, but also because of inferior text-image correspondence and poor image quality. Here we propose DATID-3D, a domain adaptation method tailored for 3D generative models using text-to-image diffusion models that can synthesize diverse images per text prompt without collecting additional images and camera information for the target domain. Unlike 3D extensions of prior text-guided domain adaptation methods, our novel pipeline was able to fine-tune the state-of-the-art 3D generator of the source domain to synthesize high resolution, multi-view consistent images in text-guided targeted domains without additional data, outperforming the existing text-guided domain adaptation methods in diversity and text-image correspondence. Furthermore, we propose and demonstrate diverse 3D image manipulations such as one-shot instance-selected adaptation and single-view manipulated 3D reconstruction to fully enjoy diversity in text.
    \end{abstract}
\vspace{-1em}

\section{Introduction}
\label{sec1_introduction}

Recently, 3D generative models~\cite{chan2021pi, gadelha20173d, hao2021gancraft, henzler2019escaping, liao2020towards, nguyen2019hologan, nguyen2020blockgan, niemeyer2021giraffe, schwarz2020graf, shi2021lifting, szabo2019unsupervised, wu2016learning, zhu2018visual, gu2022stylenerf, zhou2021cips, chan2022efficient } have been developed to extend 
2D generative models 
for multi-view consistent and explicitly pose-controlled image synthesis.
Especially,
some of them~\cite{gu2022stylenerf, zhou2021cips, chan2022efficient} combined 2D CNN generators like StyleGAN2~\cite{karras2020analyzing} with 3D inductive bias from the neural rendering~\cite{mildenhall2020nerf}, enabling efficient synthesis of high-resolution photorealistic images with remarkable view consistency and detailed 3D shapes.
These 3D generative models can be trained with single-view images and then can sample infinite 3D images in real-time, 
while 3D scene representation as neural implicit fields using NeRF~\cite{mildenhall2020nerf} and its variants~\cite{zhang2020nerf++, gafni2021dynamic, deng2022depth, takikawa2021neural, lindell2021autoint, lin2021barf, boss2021nerd, rebain2021derf, barron2021mip, reiser2021kilonerf, park2021nerfies, srinivasan2021nerv, peng2021neural, yariv2021volume, garbin2021fastnerf, hedman2021baking, chen2021mvsnerf, martin2021nerf, yu2021pixelnerf, liu2020neural, yu2021plenoctrees, pumarola2021d} require multi-view images and training 
for each scene.

Training these state-of-the-art 3D generative models is challenging  
because it requires not only a large set of images but also the information on 
the camera pose distribution of those images.
This requirement, unfortunately, has  
restricted 
these 3D models to the handful  
domains where camera parameters are annotated (ShapeNetCar~\cite{chang2015shapenet, sitzmann2019scene}) or off-the-shelf pose extractors are available (FFHQ~\cite{karras2019style}, AFHQ~\cite{karras2021alias, choi2020stargan}). 
StyleNeRF~\cite{gu2022stylenerf} assumed the camera pose distribution as either Gaussian or uniform, but 
this assumption is valid only for a few pre-processed datasets.
Transfer learning methods for 2D generative models~\cite{ojha2021few, mo2020freeze, robb2020few, pinkney2020resolution, noguchi2019image, wang2020minegan, li2020few, tseng2021regularizing } with small dataset can 
widen the scope of 3D models potentially for multiple domains, but
are also limited to a handful of domains 
with similar camera pose distribution as
the source domain in practice.

Text-guided domain adaptation methods~\cite{gal2021stylegan, alanov2022hyperdomainnet} have been developed for 2D generative models as a promising approach to bypass the additional data curation issue for the target domain. Leveraging the 
CLIP (Contrastive Language-Image Pre-training) models~\cite{radford2021learning} pre-trained on a large number of image-text pairs 
with non-adversarial fine-tuning strategies, these methods perform text-driven domain adaptation.
However, one drawback of them 
is the catastrophic loss of diversity inherent in a text prompt due to the deterministic embedding of the CLIP text encoder
so that the sample diversity of the source domain 2D generative model is not preserved in the target domain 2D generative models.

\begin{figure}[!t]
    \centering
    \includegraphics[width=0.9\linewidth]{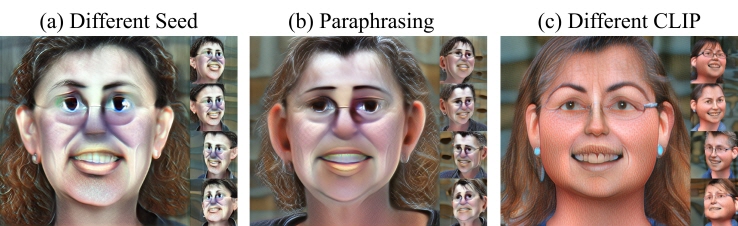}
    \vspace{-0.8em}
    \caption{Existing text-guided domain adaptation~\cite{gal2021stylegan, alanov2022hyperdomainnet} did not preserve the diversity in the source domain for the target domain.}
    \vspace{-1.5em}
    \label{fig2_clip_based_da}
\end{figure}

We confirmed this diversity loss with experiments. A text prompt ``a photo of a 3D render of a face in Pixar style'' should include lots of different  
characters' styles in Pixar films such as Toy Story, Incredible, etc.
However, CLIP-guided adapted generator can only synthesize samples with alike 
styles as illustrated in Figure~\ref{fig1_overview} (see StyleGAN-NADA$^*$).
Thus, we confirmed that  
naive extensions  
of these for 3D generative models show inferior text-image correspondence and poor quality of generated images in diversity.
Optimizing with one text embedding yielded 
almost similar results even with different training seeds as shown in Figure~\ref{fig2_clip_based_da}(a).
Paraphrasing the text for obtaining different CLIP embeddings was also trained, but it also did not yield
that many different results as illustrated in Figure~\ref{fig2_clip_based_da}(b). Using different 
CLIP encoders for a single text as in Figure~\ref{fig2_clip_based_da}(c) did provide different samples, but it was not an option in general since 
only a few CLIP encoders have been released, and retraining them requires massive servers in practice.

We propose a novel DATID-3D, a method of Domain Adaptation using Text-to-Image Diffusion tailored for 3D-aware Generative Models.
Recent progress in text-to-image diffusion models enables to synthesize diverse high-quality images from one text prompt~\cite{rombach2022high, ramesh2022hierarchical, saharia2022photorealistic }.
We first leverage them 
to convert the samples from the pre-trained 3D generator into diverse pose-aware target images. 
Then, the target images are rectified through our novel CLIP and pose reconstruction-based filtering process. 
Using these filtered target images, 3D domain adaptation is performed while preserving diversity in the text as well as multi-view consistency.
We apply our novel pipeline to the EG3D~\cite{chan2022efficient}, a state-of-the-art 3D generator, enabling the synthesis of high-resolution multi-view consistent images in text-guided target domains as illustrated in Figure~\ref{fig1_overview}, without collecting additional images with camera information for the target domains.
Our results demonstrate superior quality, diversity, and high text-image correspondence in qualitative comparison, KID, and human evaluation compared to those 
of existing 2D text-guided domain adaptation methods for 
the 3D generative models.
Furthermore, we propose one-shot instance-selected adaptation and single-view manipulated 3D reconstruction to fully enjoy diversity in the text by extending useful 2D applications of generative models.


\section{Related Works}
\label{sec2_related_works}

\subsection{3D generative models}

Recent advances in 3D generative models have achieved 
multi-view consistent and explicitly pose-controlled image synthesis.
Mesh-based~\cite{liao2020towards, szabo2019unsupervised}, voxel-based~\cite{gadelha20173d, henzler2019escaping, nguyen2019hologan, nguyen2020blockgan, zhu2018visual, wu2016learning}, block-based and fully implicit representation-based~\cite{chan2021pi, nguyen2019hologan,niemeyer2021giraffe, schwarz2020graf } 3D generative models have been 
proposed, but suffer from low image quality, view inconsistency, and inefficiency.
Recently, 
efficient models~\cite{gu2022stylenerf, zhou2021cips, chan2022efficient} have been developed to combine the state-of-the art 2D CNN generator (\textit{e.g.} StyleGAN2~\cite{karras2020analyzing}) with neural rendering~\cite{mildenhall2020nerf}.
Especially, EG3D utilizes tri-plane hybrid representation and poses conditioned dual discrimination 
to generate images with the state-of-the-art quality, view-consistency  
and 3D shapes in real-time.
Such 3D generative models can be trained using single-view images and then can sample infinite 3D images in real-time  
whereas 3D scene representation as neural implicit fields using Neural Radiance Field (NeRF)~\cite{mildenhall2020nerf} and its variants~\cite{zhang2020nerf++, gafni2021dynamic, deng2022depth, takikawa2021neural, lindell2021autoint, lin2021barf, boss2021nerd, rebain2021derf, barron2021mip, reiser2021kilonerf, park2021nerfies, srinivasan2021nerv, peng2021neural, yariv2021volume, garbin2021fastnerf, hedman2021baking, chen2021mvsnerf, martin2021nerf, yu2021pixelnerf, liu2020neural, yu2021plenoctrees, pumarola2021d} requires multi-view images and training time for each scene.

Training recent 3D generative models is more difficult than training 2D generative models since it requires 
not only a large number of images but also the information on  
the camera parameter distribution of those images.
To broadly leverage 
the state-of-the-art 3D generative models to cover wider domains, we propose a method of text-guided domain adaptation without  
additional images for the target 
domain and construct our 
novel pipeline so that the EG3D, a state-of-the-art 3D generator, can be fine-tuned to perform 
the synthesis of high-resolution multi-view consistent images in text-guided targeted domains.

\subsection{Text-guided domain adaptation using CLIP}

CLIP~\cite{radford2021learning} is composed of the image encoder $E_I^{{C}}$ and the text encoder $E_T^{{C}}$ 
that translate their inputs into vectors in a shared multi-modal CLIP space. 
StyleGAN-NADA~\cite{gal2021stylegan} fine-tunes a pre-trained StyleGAN2 $G^{\theta}$ ~\cite{karras2020analyzing} to shift the domain towards a target domain using  
a simple textual prompt guided by directional CLIP loss as follows: 
\small
\begin{align}
    \mathcal{L}^{\theta}_{\text{direction}}\left(\xb^{\text{gen}},{y^{\text{tar}}}; \xb^{\text{src}},{y^{\text{src}}}\right): = 1 - \frac{\langle \Delta I, \Delta T\rangle}{\|\Delta I\|\|\Delta T\|},
\end{align}
\normalsize
where \( \Delta I =  E_I^{{C}}(\xb^{\text{gen}}) - E_I^{{C}}(\xb^{\text{src}}), \Delta T =  E_T^{{C}}{(y^{\text{tar}}}) - E_T^{{C}}({y^{\text{src}}}) \).
Here, the CLIP space direction between the source and target images $\Delta I$
and the direction between the 
source and target text descriptions $\Delta T$ are encouraged to align. 
HyperDomainNet~\cite{alanov2022hyperdomainnet} additionally proposes a domain-modulation technique to reduce the number of trainable parameters and the in-domain angle consistency loss to avoid mode collapse. 

A critical drawback of  
these methods are that diversity inherent in a text prompt is catastrophically 
lost, resulting in alike 
samples to represent only one instance per text prompt due to the deterministic embedding of the CLIP encoder  $E_T^{C}(\xb)$. Moreover, naive extensions 
of these methods to 3D models exhibit 
inferior text-image correspondence and poor image quality. 
Our proposed DATID-3D aims to achieve 
superior quality, diversity, and high text-image correspondence to  
existing 2D text-guided domain adaptation methods for  
3D generative models qualitatively and quantitatively.

\subsection{Text-guided diffusion models}

Diffusion models
have achieved great success in image generation~\cite{ho2020denoising,song2020denoising,song2020score,jolicoeur2020adversarial,dhariwal2021diffusion}.
Recently, these models have been extended to image-text multi-modal settings, showing promising results~\cite{rombach2022high, ramesh2022hierarchical, saharia2022photorealistic, kim2022diffusionclip, avrahami2022blended}. 
Especially, text-to-image diffusion models trained on billions of image-text pairs~\cite{rombach2022high, ramesh2022hierarchical, saharia2022photorealistic} enables to synthesize outstanding quality of diverse 2D images with one target text prompt through the stochastic generation process.
Furthermore, recent progress~\cite{ruiz2022dreambooth, gal2022image} enables the synthesis of text-guided novel scenes for a given subject using only a few images.
 
In the meanwhile, the 
text-guided diffusion models for 3D scenarios are underexplored.
A concurrent work, DreamFusion~\cite{poole2022dreamfusion}, performs text-to-3D synthesis by optimizing a randomly-initialized NeRF guided by Imagen, a huge text-to-image diffusion model, that is not publicly available. 
Their results are impressive but tend to be blurry and lack fine details. Moreover, it requires a long time to generate one scene due to the  
an inherent drawback of NeRF-like methods.
Also, if Stable diffusion~\cite{rombach2022high}, a publicly available lightweight text-to-image diffusion model, is used, DreamFusion generation, but 
often failed to reconstruct 3D scenes.
In contrast, our proposed method enables the real-time synthesis of diverse high-resolution samples once trained, even with relatively small-scale, efficient diffusion models.

\begin{figure*}[!ht]
    \centering
    \includegraphics[width=0.9\linewidth]{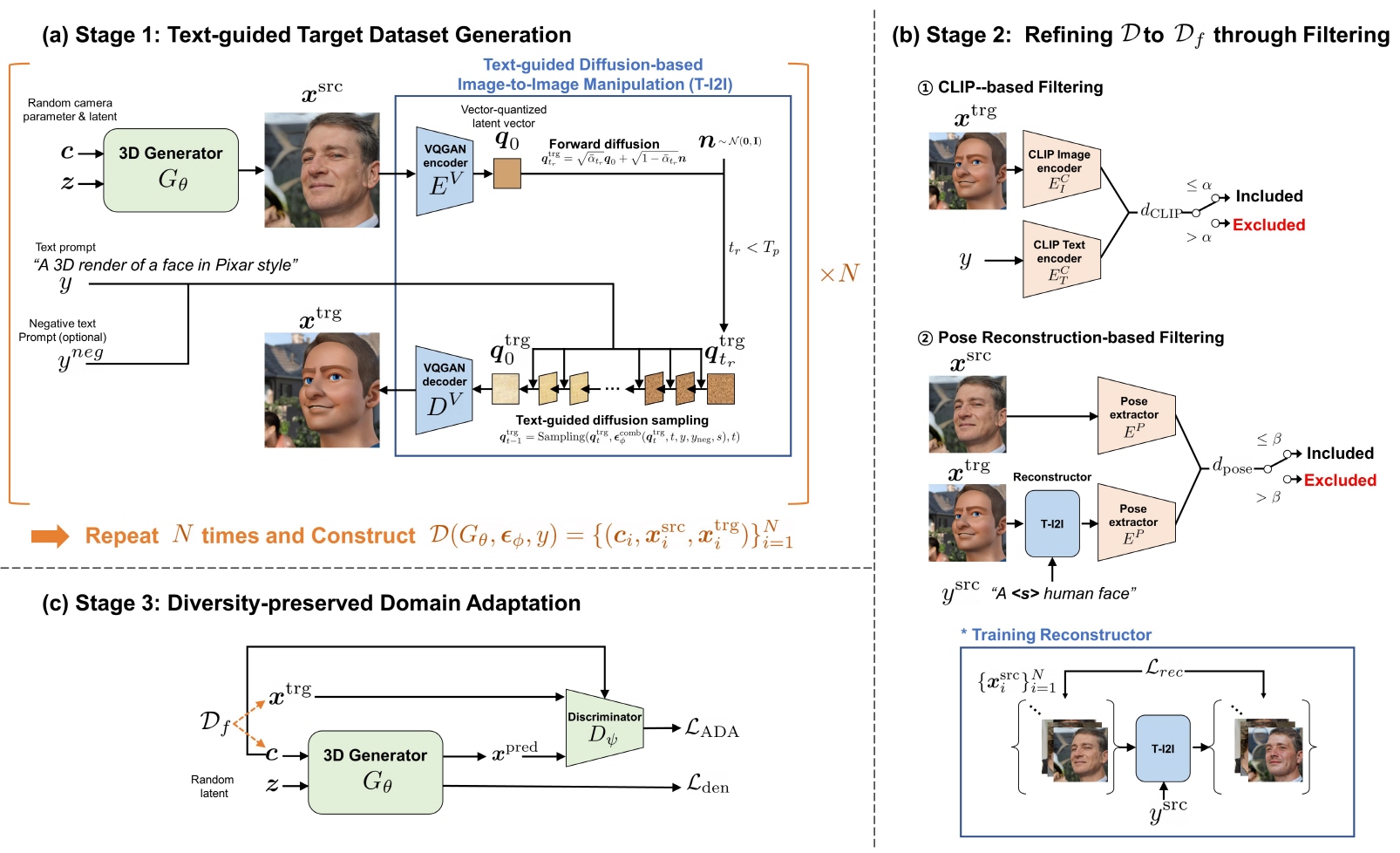}
   \vspace{-0.8em}
    \caption{Overview of DATID-3D. We construct target dataset using the pre-trained text-to-image diffusion models, followed by refining the dataset through filtering process. Finally, we fine-tune our models using adversarial loss and density regularization.    }
   \vspace{-1em}
    \label{fig3_method}
\end{figure*}

\begin{figure}[!htb]
    \centering
    \includegraphics[width=0.8\linewidth]{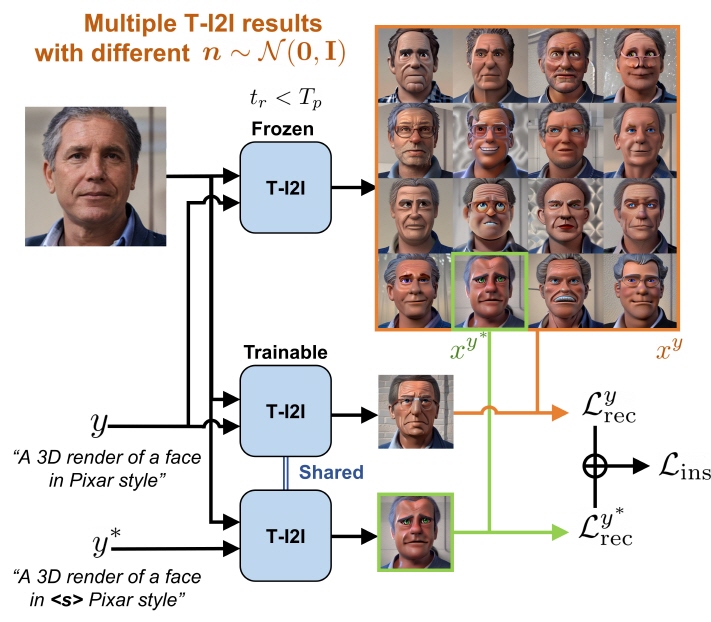}
    \vspace{-0.8em}
    \caption{One-shot fine-tuning of text-to-image diffusion models for instance-selected domain adaptation. Resulting text-to-image diffusion models are applied to the Stage 1 in Figure~\ref{fig3_method}(a). }
    \vspace{-1em} 
    \label{fig4_instance_chosen_da}
\end{figure}

\section{DATID-3D}
\label{sec3_method}

We aim to transfer EG3D~\cite{chan2022efficient}, a state-of-the-art 3D generator $G_{\theta}$ trained on a source domain, to a new target domain specified by a text prompt $y$ while preserving multi-view consistency and diversity in text.
We employ a pre-trained text-to-image diffusion model $\epsilonb_{\phi}$ as a source of supervision, but no additional image for the target domain is used. 
Firstly, we use our novel pipeline to construct a target dataset $\mathcal{D}(G_\theta, \epsilonb_{\phi}, y) = \{( \cb_i, \xb_i^{\text{src}}, \xb_i^{\text{trg}})\}_{i=1}^{N}$ that consists of random latent vectors, camera parameters and corresponding target images in a text-guided domain as illustrated in Figure~\ref{fig3_method}(a).
Secondly, we refine the dataset to obtain $\mathcal{D}_{f}$ through our CLIP and pose reconstruction-based filtering process for improved image-text correspondence and pose-consistency, respectively, as shown in Figure~\ref{fig3_method}(b).
Lastly, with the rectified dataset, we fine-tune the generator $G_{\theta}$ with 
adversarial and density-regularization losses to preserve diversity and multi-view consistency as presented in Figure~\ref{fig3_method}(c). 
In addition, we propose a one-shot instance-selected adaptation to let users fully enjoy diversity in the text as illustrated in Figure~\ref{fig4_instance_chosen_da}.

\subsection{Text-guided target dataset generation}

Here, we generate a source image $\xb^{\text{src}} = G_{\theta}(\zb, \cb)$ with random latent vector $\zb \in \mathcal{Z}$ and camera parameter $\cb \in \mathcal{C}$. 
Then, we manipulate the source image $\xb^{\text{src}}$ to yield the target image $\xb^{\text{trg}}$ guided by a text prompt $\y$ using the ideas in~\cite{meng2021sdedit} using a text-to-image diffusion model $\epsilonb_{\phi}$, constructing a set of $(\cb, \xb^{\text{src}}, \xb^{\text{trg}})$. 
Stable diffusion~\cite{rombach2022high} is selected and employed in this work since 
a latent-based model 
is lightweight and publicly available while others~\cite{ramesh2022hierarchical,saharia2022photorealistic} are not. 
$\xb^{\text{src}}$ is encoded to the latent representation $\qb_0=E^{V}(\xb^{\text{src}})$ using VQGAN~\cite{esser2021taming} encoder $E^{V}$. 
Then, the latent is perturbed through the stochastic forward DDPM (Denoising Diffusion Probabilistic Models) process~\cite{ho2020denoising} with noise schedule parameters $\{\bar\alpha_{t}\}_{t=1}^{T}$ until the return step $t_r < T_p$, where $T$ is the number of total diffusion steps used in training  
and $T_p$ is the pose-consistency step which is the last diffusion step when the pose information is preserved as follows:
 \begin{align}
    \qb^{\text{trg}}_{t_r} = \sqrt{\bar\alpha_{t_r}}\qb_{0}  + \sqrt{1 - \bar\alpha_{t_r}}\nb, \  \text{where} \ \nb \sim \mathcal{N} (\mathbf{0,I}).
\end{align}
We set $T=1000$ following 
\cite{ho2020denoising,song2020denoising, dhariwal2021diffusion} and $T_p=850$ based on the experimental results.
Then, we generate the manipulated target latent $\qb_{0}^{trg}$ from the perturbed latent $\qb_{t_r}^{\text{trg}}$ through text-guided sampling process as follows:
\small
\begin{align}
    \qb^{\text{trg}}_{t-1} = \text{Sampling}(\qb^{\text{trg}}_{t}, \epsilonb^{\text{comb}}_\phi(\qb^{\text{trg}}_{t},{t}, y, y_{\text{neg}}, s), t)
\end{align}
\normalsize
where $s$ is a guidance scale parameter 
that controls the scale of gradients from a target prompt $y$ and a negative prompt $y_{\text{neg}}$ and 
\(     \epsilonb^{\text{comb}}_\phi = s\epsilonb_\phi(\qb^{\text{trg}}_{t},{t}, y) + (1-s)\epsilonb_\phi(\qb^{\text{trg}}_{t},{t}, y_{\text{neg}}). \)
Any sampling method such as DDPM~\cite{ho2020denoising}, DDIM~\cite{song2020denoising}, PLMS~\cite{liu2022pseudo} can be used. 
We can specify $y_{\text{neg}}$ to prevent the manipulated image from being contaminated by an unwanted concept or can just leave it as an unconditional text token.
Then, we can obtain the target image $\xb^{\text{trg}}=D^{V}(\qb^{\text{trg}}_0)$ using the VQGAN decoder that represents one of the diverse concepts inherent in the text prompt.
By repeating this process $N$ times, we can get a target dataset $\mathcal{D}(G_\theta, \epsilonb_{\phi}, y) = \{( \cb_i, \xb_i^{\text{src}}, \xb_i^{\text{trg}})\}_{i=1}^{N}$. 

\subsection{CLIP and pose reconstruction-based filtering}
We found that the raw target dataset $\mathcal{D}$ may sometimes include target images that may not correspond to the target text or that the camera pose is changed during the stochastic process. 
To resolve this issue, we propose CLIP-based and pose reconstruction-based filtering processes for enhanced image-text correspondence and pose consistency. 

\paragraph{CLIP-based filtering.}
A CLIP distance score ${d}_{\text{CLIP}}$ is the cosine distance in the CLIP space between a potential target image $\xb^{\text{trg}}$ and a text prompt $y$ and if 
\begin{equation}
    {d}_{\text{CLIP}}(\xb^{\text{trg}}, {y}) = 1 - \frac{\langle E_I^{{C}}(\xb^{\text{trg}}), E_T^{{C}}(y) \rangle}{\|E_I^{{C}}(\xb^{\text{trg}})\|\|E_T^{{C}}(y)\|} > \alpha
    \label{eq_clip_distance}
\end{equation}
where $\alpha$ is a chosen threshold, then $\xb^{\text{trg}}$ is removed from $\mathcal{D}$.

\paragraph{Pose reconstruction-based filtering.}
A pose difference score ${d}_{\text{pose}}$ is the $l_2$ distance between the poses extracted from $\xb$ and $\xb'$ using an off-the-shelf pose extractor $E^{\text{P}}$: 
\begin{equation}
    {d}_{\text{pose}}(\xb, \xb') = \|E^{\text{P}}(\xb) - E^{\text{P}}(\xb')\|_2^2. 
    \label{eq_pose_difference}
\end{equation}
To calculate pose difference, we leverage the universal Reconstructor that converts the target images from any shifted domain to the source domain (\textit{e.g.}, human face) where the pose extractor is available.
We fine-tuned the pre-trained text-to-image diffusion models to generate the source domain images $\{\xb_i^{\text{src*}}\}_{i=1}^{N}$ using the following diffusion-based reconstruction loss $\mathcal{L}^{\phi}_{\text{rec}}$: 
\small
\begin{align}
\mathbb{E}_{\qb \in \{E^V(\xb_i^{\text{src}})\}^{N}_{i=1}, \epsilonb \sim \mathcal{N}(0,1), t}[\|\epsilonb-\epsilonb_\phi(\zb_t, t, y^{\text{src}}))\|_2^2]
\end{align}
\normalsize
where $y^{\text{src}}$ is a text prompt representing the source domain with a specifier word <s> (\textit{e.g.}, 
``A photo of <s> face'' in FFHQ~\cite{karras2019style}). 
The Reconstructor can be re-used for any target domain if the source domain is the same. 
Using the Reconstructor, we first convert the target image $\xb^{\text{trg}}$ to a reconstructed image $\xb^{\text{rec}}$.
Then, if ${d}_{\text{pose}}(\xb^{\text{rec}}, \xb^{\text{src}}) > \beta$ ($\beta$ is a threshold), then $\xb^{\text{trg}}$ is excluded from $\mathcal{D}$ and another target image with same $(\zb, \cb)$ is generated to supplement $\mathcal{D}$.

\subsection{Diversity-preserved domain adaptation}

With the filtered dataset $\mathcal{D}_f = \{(\cb_i, \xb_i^{\text{src}}, \xb_i^{\text{trg}})\}_{i=1}^{N}$, we can perform either non-adversarial fine-tuning using CLIP-based loss like  
StyleGAN-NADA~\cite{radford2021learning} and HyperDomainNet~\cite{alanov2022hyperdomainnet} or adversarial fine-tuning like 
StyleGAN-ADA~\cite{karras2020training}.
As analyzed 
Section~\ref{subsec_exp_ablation}, we found that non-adversarial fine-tuning makes the generator lose diversity for the target text prompt and generates the samples representing one averaged concept among diverse concepts as well as showing sub-optimal quality because the cosine similarity loss can be saturated near the optimal point. 
In contrast, we found that adversarial fine-tuning preserves diverse concepts in the text. 
Our total loss to train pose-conditioned generator $G_\theta$ and discriminator $D_\psi$ consists of ADA loss $\mathcal{L}_{\text{ADA}}$ and density regularization loss $\mathcal{L}_{\text{den}}$, which were used in EG3D~\cite{chan2022efficient}. 
\paragraph{ADA loss.}  
ADA loss $\mathcal{L}_{\text{ADA}}$~\cite{karras2020training}  
is an adversarial loss with adaptive dual augmentation and R1 regularization as follows:
\small
\begin{align}
    &\mathcal{L}^{\theta,\psi}_{\text{ADA}}=\mathbb{E}_{\zb \sim \mathcal{Z}, \cb \sim \mathcal{C}} [f(D_\psi(A(G_\theta(\zb, \cb)), \cb)]  \\
    &+\mathbb{E}_{(\cb, \xb^{\text{trg}}) \in \mathcal{D}_f}[f(-D_{\psi}(A(\xb^{\text{trg}}), \cb)+\lambda\|\nabla D_{\psi}(A(\xb^{\text{trg}}), \cb)\|^2)] \nonumber
\end{align}
\normalsize
where $f(u)=-\log (1+\exp (-u))$, $A$ is a stochastic non-leaking augmentation operator with probability $p$. 
For more detailed information and algorithms of our pipelines, see the supplementary Section~\textred{B}.

\paragraph{Density regularization loss.}
Additionally, we use density regularization, which has been effective in reducing the occurrence of unwanted other shape distortions by promoting the smoothness of the density field~\cite{chan2022efficient}.
We randomly select points $v$ from the volume $\mathcal{V}$ for each rendered scene and also select additional perturbed points that have been slightly distorted by Gaussian noise $\delta v$. 
Then, we calculate the L1 loss between the predicted densities as follows:
\begin{align}
    \mathcal{L}^{\theta}_{\text{den}}= \mathbb{E}_{v\in\mathcal{V}}[\|\sigma_\theta(v) - \sigma_\theta(v+\delta v)\|].
\end{align}

\subsection{Instance-selected domain adaptation}

Our method yields a domain-shifted 3D generator to synthesize samples to represent diverse concepts in the text.
However, consider a scenario to pick up one of those diverse concepts and adapt the generator to this specific concept. 
Is it possible to manipulate our single 2D image guided by text and lift it to 3D with multiple versions?

To enable these applications and help users fully enjoy diversity in text, we propose a one-shot instance-selected adaptation.  
We first manipulate a source domain image in $N_d$ multiple versions from a single text prompt as shown in Figure~\ref{fig4_instance_chosen_da}.
Then, we choose one instance among those diverse instances and fine-tune our text-to-image diffusion models.
However, unlike the prior work~\cite{ruiz2022dreambooth} to personalize the object and use it guided by the text, our goal is to inject a specific concept among many concepts, that are implicit in one text prompt, to the text-to-image diffusion model.
Also, we fine-tune the text-to-image diffusion model only with a single image by limiting the diffusion time step from 0 to the pose-consistency step $T_p$ using the following loss $\mathcal{L}^{\phi}_{\text{ins}}$: 
\small
\begin{align}
&\mathbb{E}_{\epsilonb \sim \mathcal{N}(0,1), t\in{[0,T_p]}}[\|\epsilonb-\epsilonb_\phi(E^V(\xb^{y^*}), t, y^{*}))\|_2^2] \\
&+\mathbb{E}_{\zb \in \{E^V(\xb_i^{y})\}^{N_d}_{i=1}, \epsilonb \sim \mathcal{N}(0,1), t\in{[0,T_p]}}[\|\epsilonb-\epsilonb_\phi(\zb_t, t, y))\|_2^2] \nonumber
\end{align}
\normalsize
where $y$ and $\xb^y$ are the target text prompt and manipulated image with the prompt, respectively. $y^*$ and $\xb^{y^*}$ are the target text prompt with a specifier word <s> and the selected instance image, respectively.
The next step is 
just replacing the original text-to-diffusion model with 
our specified model in text-guided target generation stage in Figure~\ref{fig3_method}(a). 

Additionally, we can now perform single-view manipulated 3D reconstruction representing our chosen concept by combining the 3D GAN inversion method with instance-selected domain adaptation.

\begin{figure*}[!ht]
    \centering
    \includegraphics[width=1\linewidth]{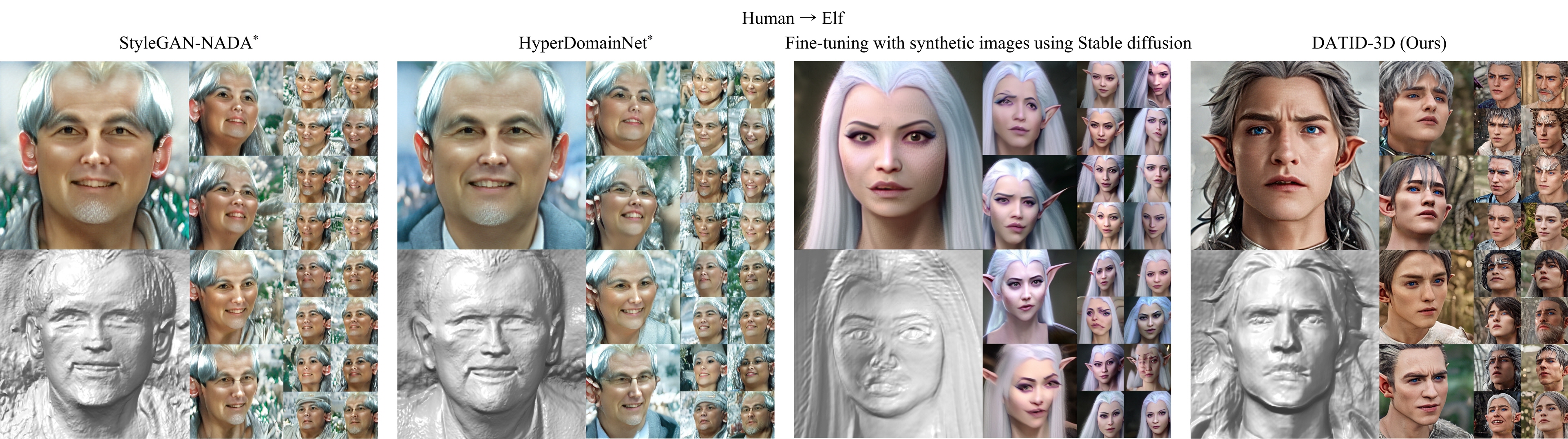}
   \vspace{-1.7em}
    \caption{Qualitative comparison with existing text-guided domain adaptation methods. The star mark (*) refers to the 3D extension of each method that is developed for 2D generative models. Our DATID-3D yielded diverse samples while other baselines did not. Naively fine-tuning 3D generators with the synthetic images using T2I diffusion resulted in losing pose-controllability and 3D shapes. For more results, see the supplementary Figure~\textred{S4}.}
   \vspace{-1em}
    \label{fig5_comparision}
\end{figure*}

\begin{figure*}[!ht]
    \centering
    \includegraphics[width=0.92\linewidth]{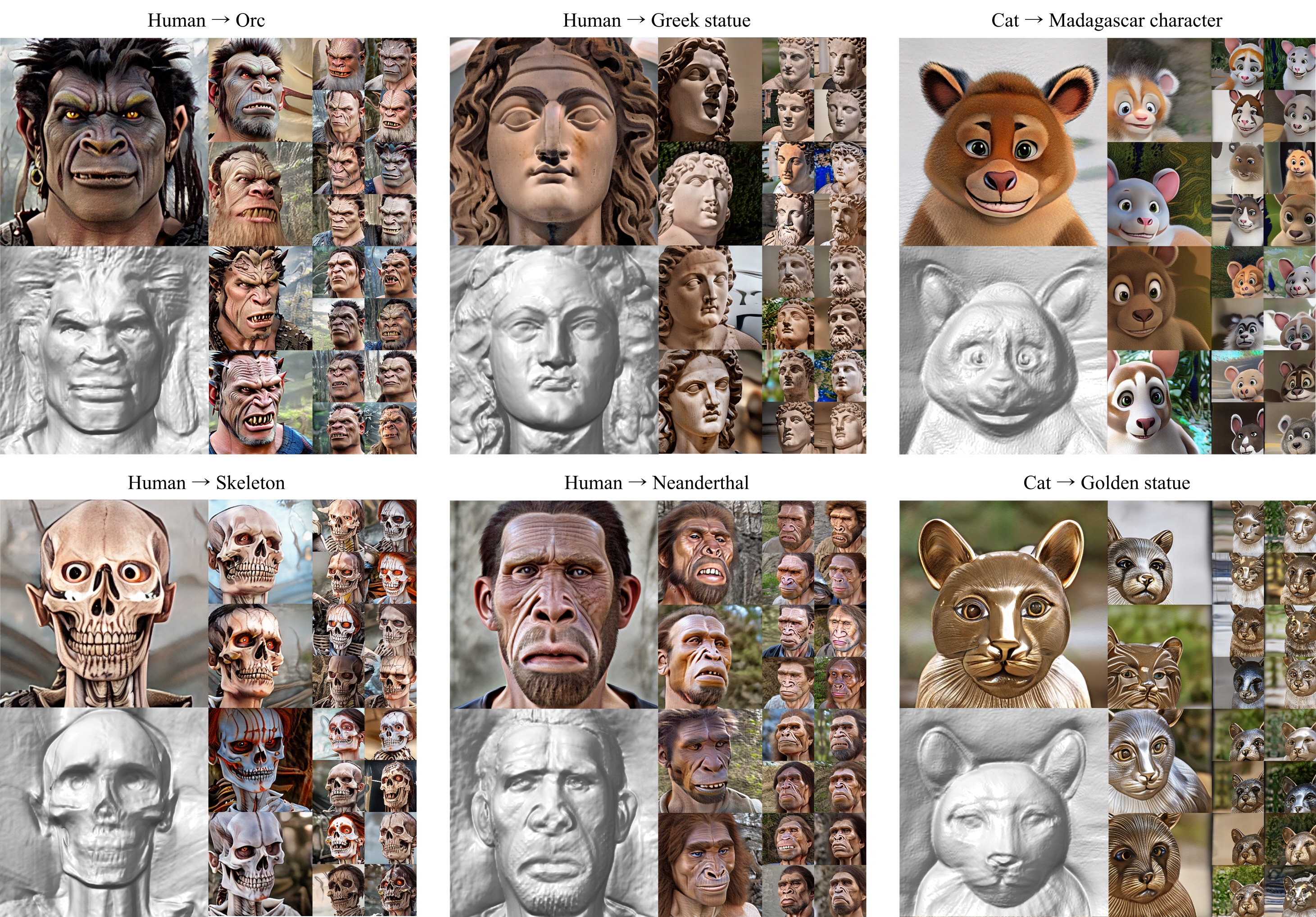}
    \vspace{-0.5em}
    \caption{Wide range of out-of-domain text-guided adaption results. We fine-tuned EG3D~\cite{chan2022efficient}, pre-trained on $512^2$ images in FFHQ~\cite{karras2019style} and AFHQ Cats~\cite{karras2021alias, choi2020stargan}, respectively, to generate diverse samples for a wide range of concepts.  For more results, see the supplementary Figure~\textred{S1}, \textred{S2} and \textred{S3}.}
    \vspace{-1.5em}
    \label{fig6_more_results}
\end{figure*}


\begin{table}[!htb]
\caption{Quantitative comparisons with the baselines in diversity, text-image correspondence and photo realism.}\label{tab1_comparision}
\centering
\begin{adjustbox}{width=0.82\linewidth}
\begin{tabular}{lccc}
\toprule
              & Text-Corr.↑    & Realism↑       & Diversity↑     \\
 \midrule
StyleGAN-NADA$^{*}$     & 2.583          & 2.550          & 2.587          \\
HyperDomainNet$^{*}$    & 2.530          & 2.520          & 2.557          \\
\textbf{Ours}           & \textbf{3.573} & \textbf{3.437} & \textbf{3.347} \\
\bottomrule
\end{tabular}
\end{adjustbox}
\end{table}

\section{Experiments}
\label{sec3_experiments}
For the experiments, we employ the publicly available lightweight Stable diffusion~\cite{rombach2022high} as our pre-trained text-to-image diffusion model.
We apply our novel pipeline to the state-of-the-art 3D generators, EG3D~\cite{chan2022efficient}, pre-trained on $512^2$ images in FFHQ~\cite{karras2019style} and AFHQ Cats~\cite{karras2021alias, choi2020stargan}, respectively. The pre-trained pose-extractor~\cite{hempel20226d} was used. 
For fine-tuning the generator, 3,000 target images were used. We set $\alpha=0.7$ and $\beta=150$.
For more detailed information about the setup of experiments, see the supplementary Section~\textred{C} and~\textred{D}.

\subsection{Evaluation}
\paragraph{Baselines.} 
To the best of our knowledge, our method is the first method of text-guided domain adaptation tailored for 3D-aware generative models.
Thus, we compare our method with CLIP-based method for 2D generative models, StyleGAN-NADA~\cite{radford2021learning} and HyperDomainNet~\cite{alanov2022hyperdomainnet}. The star mark ($^*$) refers to the extension of these method to 3D models.
To achieve this, we just add random sampling of camera parameters, followed by the random latent sampling. In StyleGAN-NADA$^*$, the directional CLIP loss is used to encourage the correspondence between the rendered 2D image with the text prompt. In addition to this, in-domain angle loss is used for HyperdomainNet$^*$.

\paragraph{Qualitative results.}
As shown in Figures~\ref{fig1_overview} and~\ref{fig5_comparision}, the generator shifted by the baseline methods fail to generate high-quality samples, preserving diversity implicit in the text prompt.
Even though in-domain angle loss in HyperDomainNet$^*$~\cite{alanov2022hyperdomainnet} is proposed to improve sample diversity, it shows similar inferior results because the fundamental issue, the deterministic text embedding of the CLIP encoder, is not resolved.
On the contrary, our DATID-3D enables the shifted generator to synthesize photorealistic and diverse images, leveraging text-to-image diffusion models and adversarial training.
In addition, we present the results of naively fine-tuning the 3D generator with synthetic images generated from random noise using Stable diffusion~\cite{rombach2022high}. However, this approach loses 3D shapes, depth, and pose-controllability, whereas our method, which uses a pose-aware target dataset, preserves 3D geometry effectively.

\paragraph{Quantitative results.}
We perform a user study to assess the perceptual quality of the produced samples. 
To quantify opinions, we requested users to rate the perceptual quality on a scale of 1 to 5, based on the following questions: (1) Do the generated samples accurately reflect the target text's semantics? (text-correspondence), (2) Are the samples realistic? (photorealism), (3) Are the samples diverse in the image group? (diversity).
We use the EG3D pre-trained on $512^2$ images in FFHQ~\cite{karras2019style} and choose four text prompts converting a human face to `Pixar', `Neanderthal', `Elf' and `Zombie' styles, respectively, for evaluation as these prompts are used in the previous work, StyleGAN-NADA~\cite{radford2021learning}.
As presented in Table~\ref{tab1_comparision}, our results demonstrate the superior quality, high diversity, and high text-image correspondence of our proposed method as compared to the baselines. 
For more results and details on the comparison, see the supplementary Section~\textred{A} and \textred{D}.

\subsection{Results of 3D out-of-domain adaptation}
We display a wide range of text-driven adaptation results through our methods in Figure~\ref{fig6_more_results}, which are applied to the generators pre-trained on FFHQ~\cite{karras2019style} or AFHQ Cats~\cite{karras2021alias, choi2020stargan}.
Our model enables the synthesis of high-resolution multi-view consistent images in various text-guided out-of-domains beyond the boundary of the trained domains, without additional images and camera information. For more results, see the supplementary Section~\textred{A}.

\begin{figure}[!tb]
    \centering
    \includegraphics[width=0.9\linewidth]{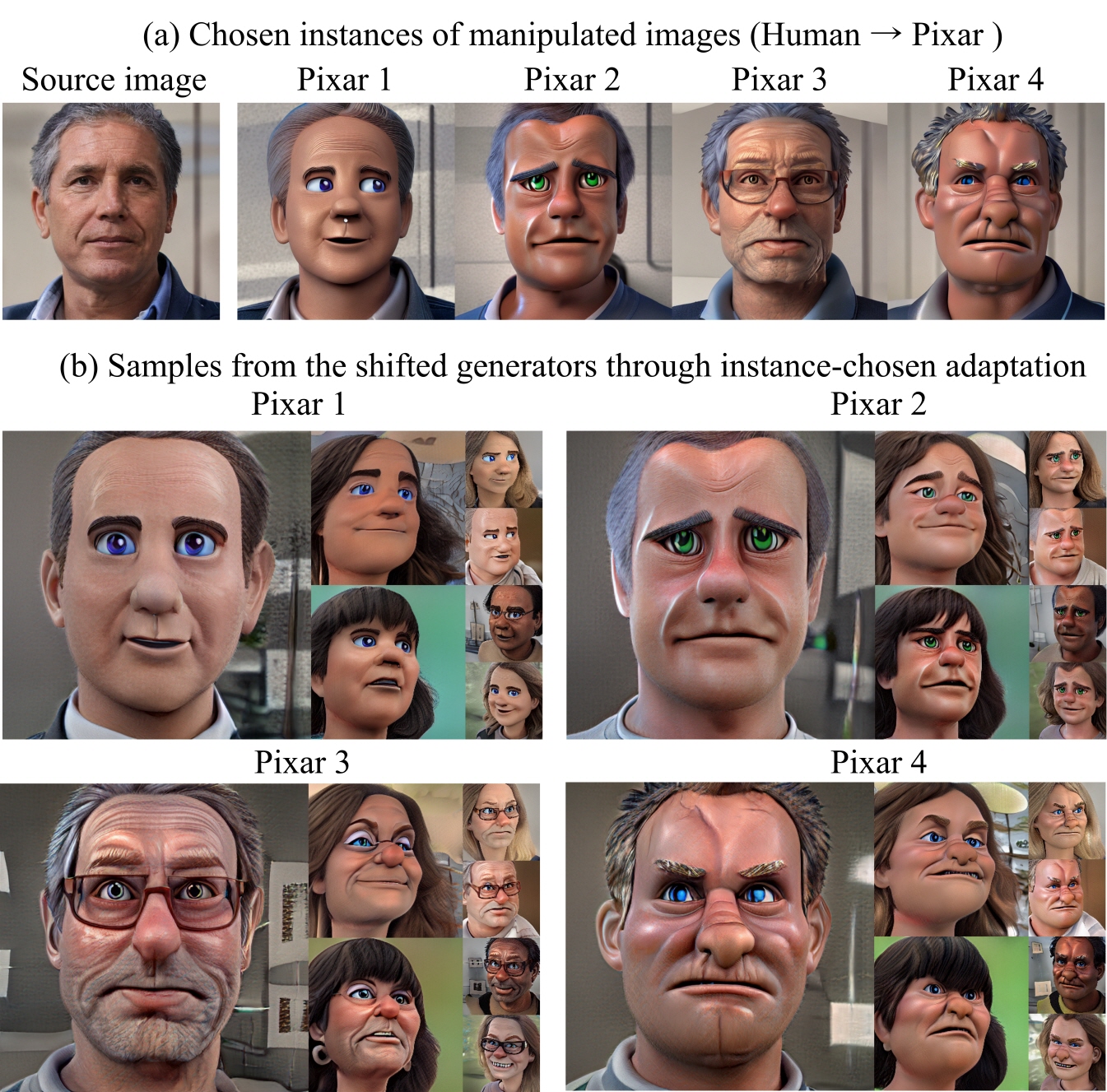}
    \vspace{-0.5em}
    \caption{Results of instance-selected domain adaptation, selecting one Pixar sample to generate more diverse samples for it.}
    \vspace{-0.5em}
    \label{fig7_style_chosen_adapt}
\end{figure}

\begin{figure}[!tb]
    \centering
    \includegraphics[width=0.95\linewidth]{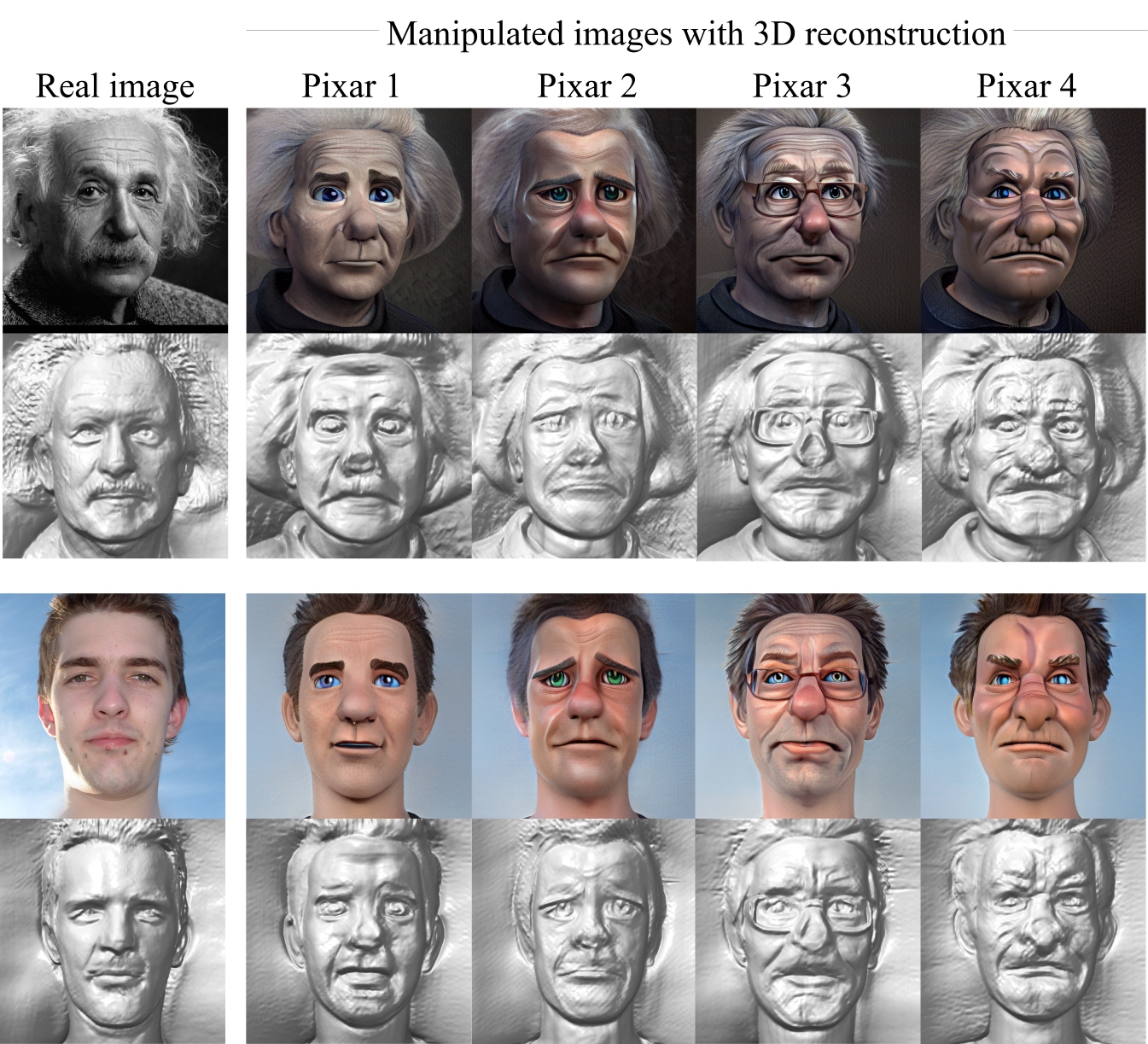}
    \vspace{-0.5em}
    \caption{Results of single-view manipulated 3D reconstruction, generating diverse 3D images on other domains with view consistency for a given single real image.}
    \vspace{-1.em}
    \label{fig8_manipulated_3d_recon}
\end{figure}

\begin{figure}[!tb]
    \centering
    \includegraphics[width=0.9\linewidth]{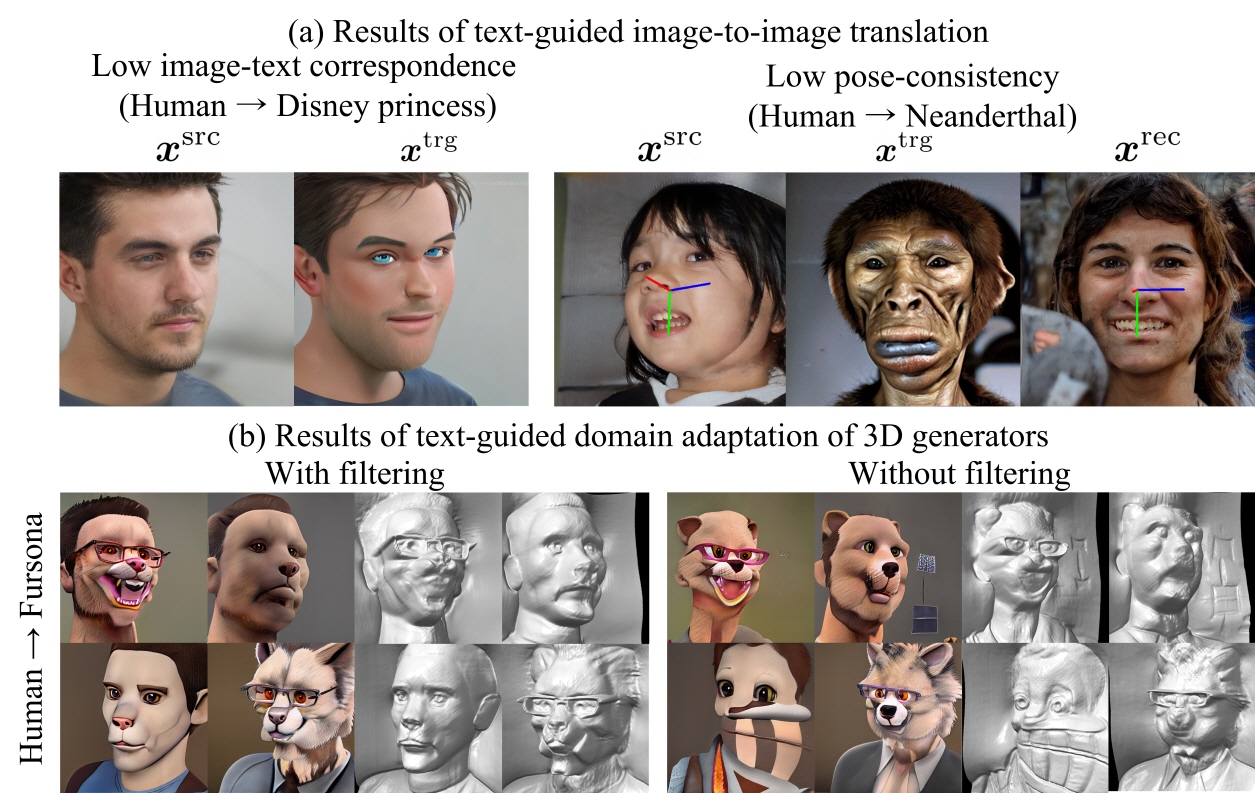}
    \vspace{-0.5em}
    \caption{Discarded cases through our filtering process (a) and results of domain adaptation of 3D generative models with and without filtering (filtering rate $= 0.529$)}
    \vspace{-0.5em}
    \label{fig9_ablation_filtering}
\end{figure}

\begin{figure}[!tb]
    \centering
    \includegraphics[width=0.9\linewidth]{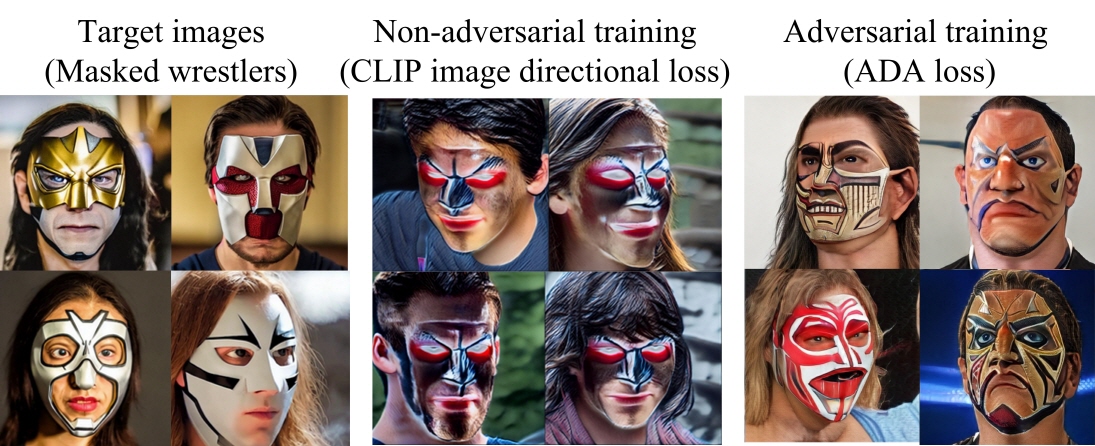}
    \vspace{-0.5em}
    \caption{Diversity preservation can be ensured not by non-adversarial training loss, but by adversarial training loss.}
    \vspace{-0.5em}
    \label{fig10_loss_ablation}
\end{figure}


\begin{figure}[!tb]
    \centering
    \includegraphics[width=0.9\linewidth]{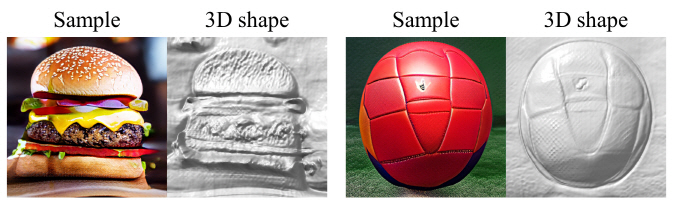}
  \vspace{-0.5em}
    \caption{Rotation-invariant objects lose 3D shapes during fine-tuning due to the lack of information about directions.}
    \vspace{-1.5em}
    \label{fig11_case_studies}
\end{figure}

\subsection{Instance-selected domain adaptation}
\label{subsec_exp_instance_selected}
In Figure~\ref{fig7_style_chosen_adapt}, we adapt our generator guided by text prompt, ``a photo of 3D render of a face in Pixar style'', in four different versions. Figure~\ref{fig7_style_chosen_adapt}(a) presents the selected 4 cases. Figure~\ref{fig7_style_chosen_adapt}(b) displays the images sampled from the generator adapted to each instance. We can further utilize these for single-view 3D manipulated reconstruction.

\subsection{Single-view 3D manipulated reconstruction}
\label{subsec_exp_single_view_manipulated_recon}
As advancements of prior 2D text-guided image manipulation~\cite{gal2021stylegan,patashnik2021styleclip, kim2022diffusionclip}, our method enables (1) lifting the text-guided manipulated images to 3D and (2) choosing one among diverse results from one text prompt.
Figure~\ref{fig8_manipulated_3d_recon} shows the results of single-view manipulated 3D reconstruction where instance-selected domain adaptation is combined with the 3D GAN inversion method~\cite{chan2022efficient}.

\subsection{Ablation studies}
\label{subsec_exp_ablation}
\paragraph{Effectiveness of filtering process.}
Frequent failure cases of image manipulation by text-to-image manipulation were observed as shown in Figure~\ref{fig9_ablation_filtering}(a).
Our filtering process improves perceptual quality and the quality of 3D shape, especially with filtering rate $>0.5$ Fig.~\ref{fig9_ablation_filtering}(b).  

\paragraph{Adversarial vs non-adversarial fine-tuning.}
We compare the results of fine-tunings using the adversarial ADA loss with those using CLIP-based non-adversarial loss for the target images.
For the non-adversarial loss, we employ the image directional CLIP loss that tries to align the direction between source and generated images with the direction between source and target images.
As illustrated in Figure~\ref{fig10_loss_ablation}, we found that non-adversarial fine-tuning makes the generator lose diversity in the target text prompt and generates the samples representing one averaged concept among diverse concepts.
Furthermore, it shows sub-optimal quality because the cosine similarity loss can be saturated near the optimal point. 
In contrast, we observed that adversarial fine-turning preserves diverse concepts in text with excellent quality.
For more results of ablation studies, see the supplementary Section~\textred{E}.

\section{Discussion and Conclusion}
\paragraph{Limitation.}
\label{subsec_exp_analysis}
We found that for the successful text-driven 3D domain adaptation, the important condition is that the target images generated in Stage 1 should preserve pose information.
However, there are some unavoidable cases to meet this condition. 
One of the cases for pose information loss is that the target object is rotation-invariant or in 2D space.
As shown in Figure~\ref{fig11_case_studies}, domain adaptations of `Human face' $\rightarrow$ `Cheeseburger' or `Human face' $\rightarrow$ `Soccer ball' failed because pose information is lost during the manipulation, reporting high pose-difference score.

Societal risks in our methods exist. 
We advise you to use our method carefully for proper purposes. Details on limitations and negative social impacts are given in the supplementary Section~\textred{F}.

\paragraph{Conclusion.}
We propose DATID-3D, a method of domain adaptation tailored for 3D generative models leveraging text-to-image diffusion models that can synthesize diverse images per text prompt.
Our novel pipeline with the 3D generator has enabled excellent quality of multi-view consistent image synthesis in text-guided domains, preserving diversity in text and outperforming the baselines qualitatively and quantitatively. Our pipeline was able to be extended for one-shot instance-selected adaptation and single-view manipulated 3D reconstruction to meet user-intended constraints.

\section*{Acknowledgements}

This work was supported by the National Research Foundation of Korea(NRF) grants funded by the Korea government(MSIT) (NRF-2022R1A4A1030579, NRF-2022M3C1A309202211), Basic Science Research Program through the National Research Foundation of Korea(NRF) funded by the Ministry of Education(NRF-2017R1D1A1B05035810) and a grant of the Korea Health Technology R\&D Project through the Korea Health Industry Development Institute (KHIDI), funded by the Ministry of Health \& Welfare, Republic of Korea (grant number: HI18C0316). Also, the authors acknowledged the financial support from the BK21 FOUR program of the Education and Research Program for Future ICT Pioneers, Seoul National University.

{\small
\bibliographystyle{ieee_fullname}
\bibliography{egbib}
}


\clearpage
\appendix

\twocolumn[{%
\renewcommand\twocolumn[1][]{#1}%

\begin{center}
\bigskip 
\bigskip 
\textbf{\Large DATID-3D: Diversity-Preserved Domain Adaptation \\ Using Text-to-Image Diffusion for 3D Generative Model \\ (Supplementary Material) \\}

\bigskip 
\bigskip 
{\large  Gwanghyun Kim$^1$ \qquad Se Young Chun$^{1,2,\dagger}$ \\
$^1$Dept. of Electrical and Computer Engineering, $^2$INMC \& IPAI\\
Seoul National University, Korea\\
{\tt\small \{gwang.kim, sychun\}@snu.ac.kr}
}
\bigskip 
\bigskip 
\maketitle
 
\end{center}%
}]
{
  \renewcommand{\thefootnote}%
    {\fnsymbol{footnote}}
  \footnotetext[1]{Corresponding author.}
}

\setcounter{equation}{0}
\setcounter{figure}{0}
\setcounter{table}{0}
\setcounter{page}{1}
\makeatletter
\renewcommand{\theequation}{S\arabic{equation}}
\renewcommand{\thefigure}{S\arabic{figure}}
\renewcommand{\thetable}{S\arabic{table}}

\begin{figure*}[!htb]
    \centering
    \vspace{-2em}
    \includegraphics[width=0.9\linewidth]{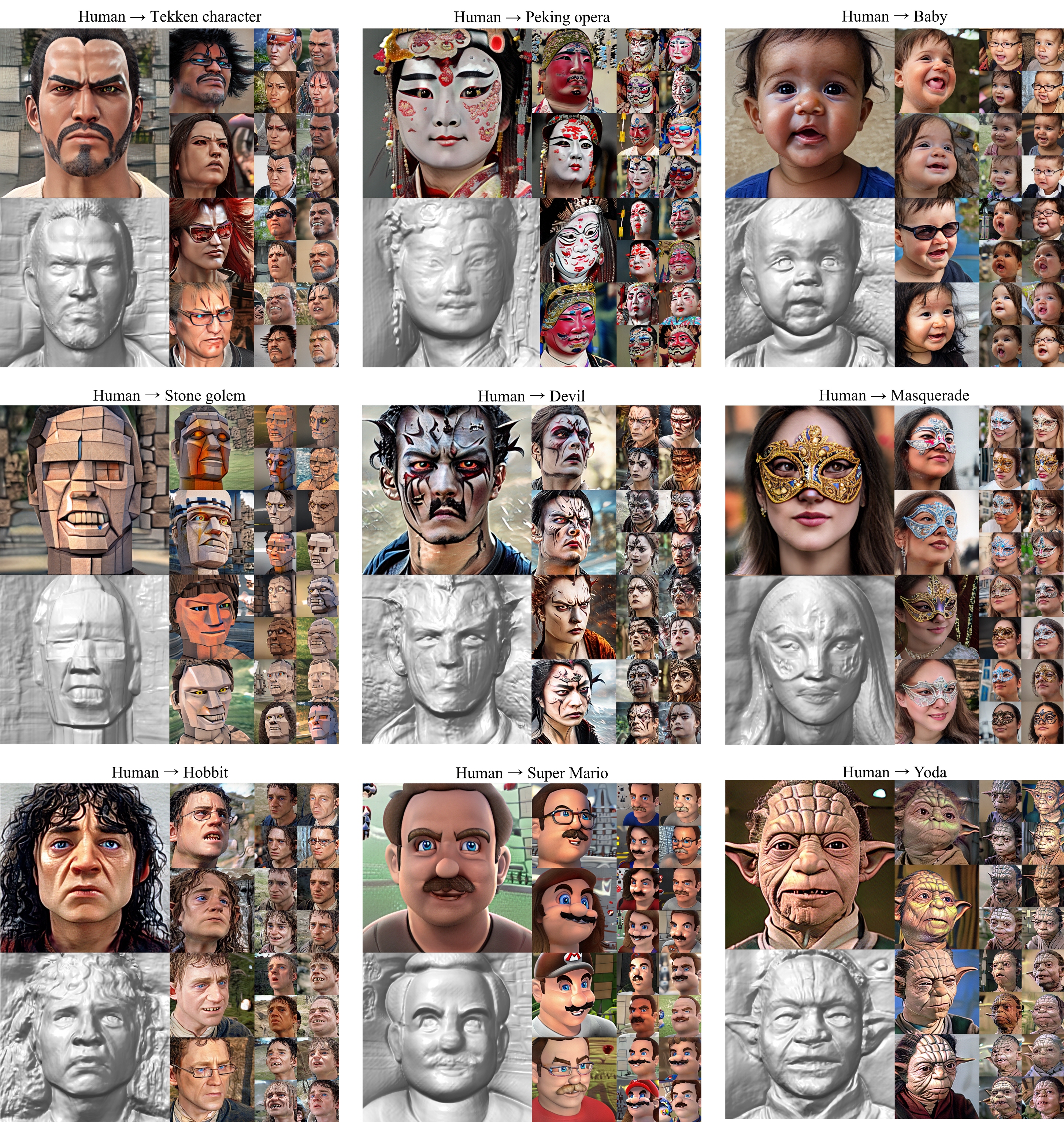}
    \caption{Variety 
    of text-guided adaption results. We fine-tuned EG3D~\cite{chan2022efficient}, pre-trained on $512^2$ images in FFHQ~\cite{karras2019style}, to generate diverse samples for a variety of concepts.}
    \vspace{-1em}
    \label{fig_supp_more_results1}
\end{figure*}

\begin{figure*}[!htb]
    \centering
    \vspace{-2em}
    \includegraphics[width=0.9\linewidth]{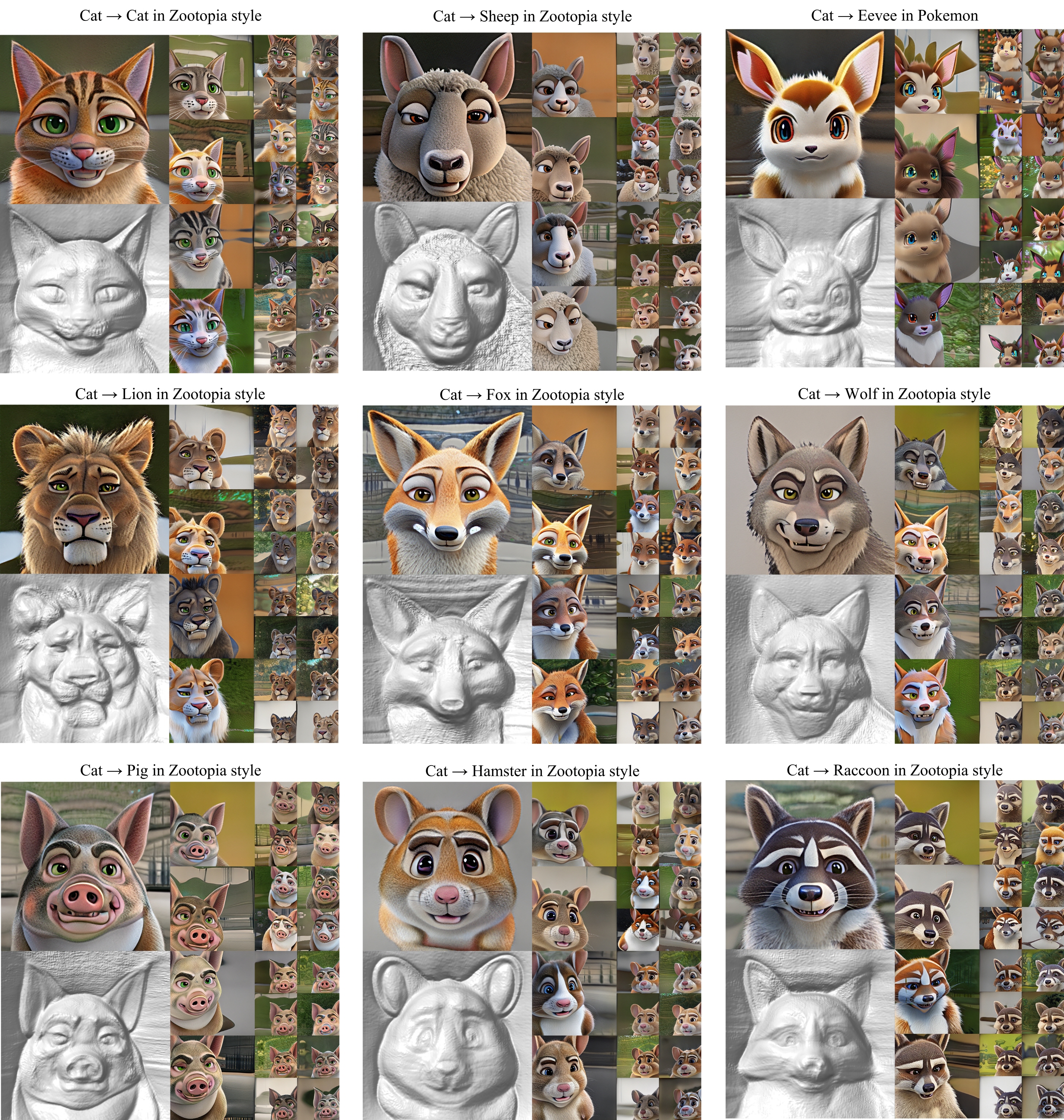}
    \caption{
    Variety of text-guided adaption results. We fine-tuned EG3D~\cite{chan2022efficient}, pre-trained on $512^2$ images in AFHQ Cats~\cite{karras2021alias, choi2020stargan} to generate diverse samples for a variety of concepts.}
    \vspace{-1em}
    \label{fig_supp_more_results2}
\end{figure*}

\begin{figure*}[!htb]
    \centering
    \vspace{-2em}
    \includegraphics[width=0.9\linewidth]{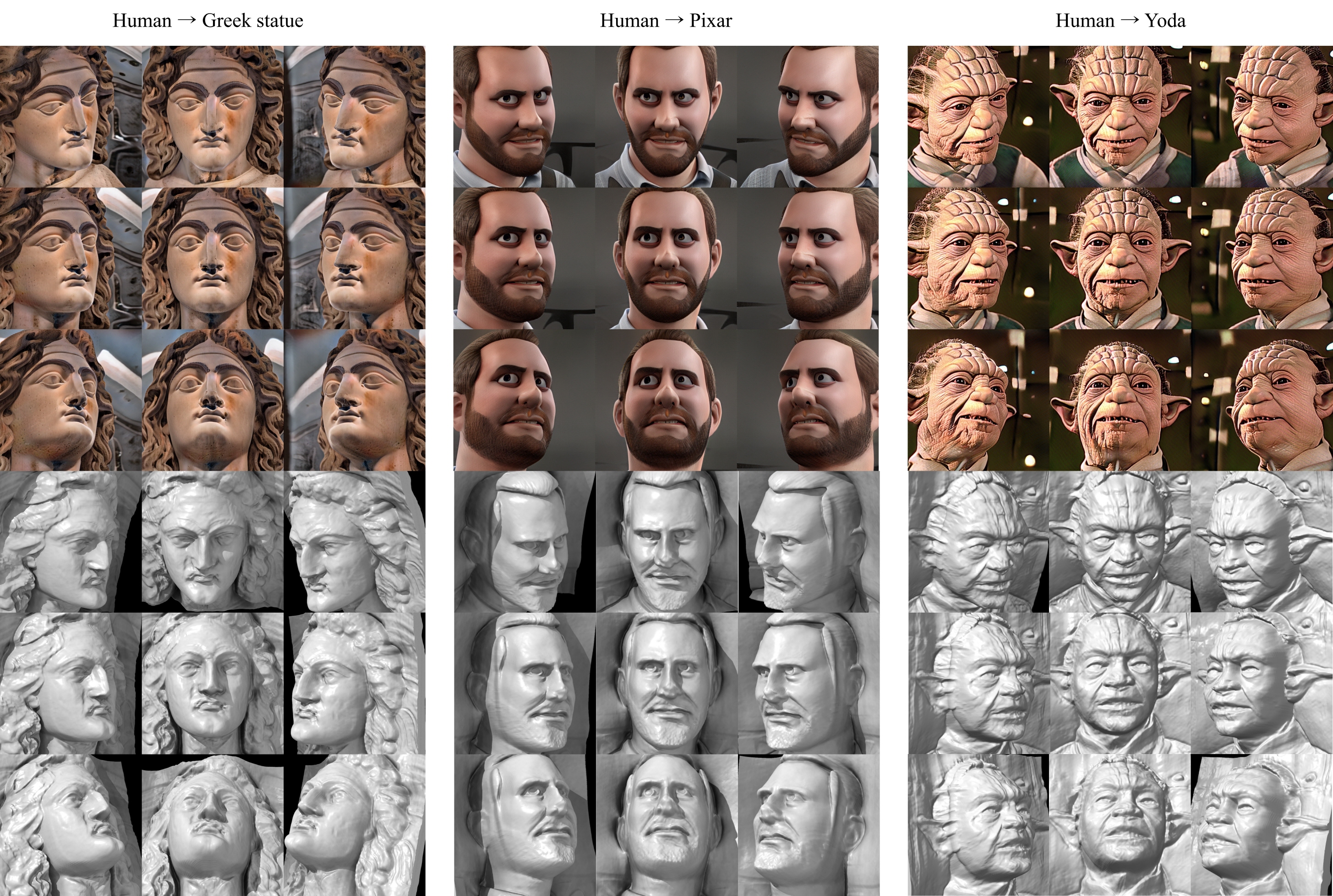}
    \caption{Pose-controlled images and 3D shapes in text-guided domain through our method. See the supplementary videos at \href{https://gwang-kim.github.io/datid_3d}{gwang-kim.github.io/datid\_3d}}
    \vspace{-1em}
    \label{fig_supp_pose_controlled}
\end{figure*}

\begin{figure*}[!htb]
    \centering
    \includegraphics[width=0.9\linewidth]{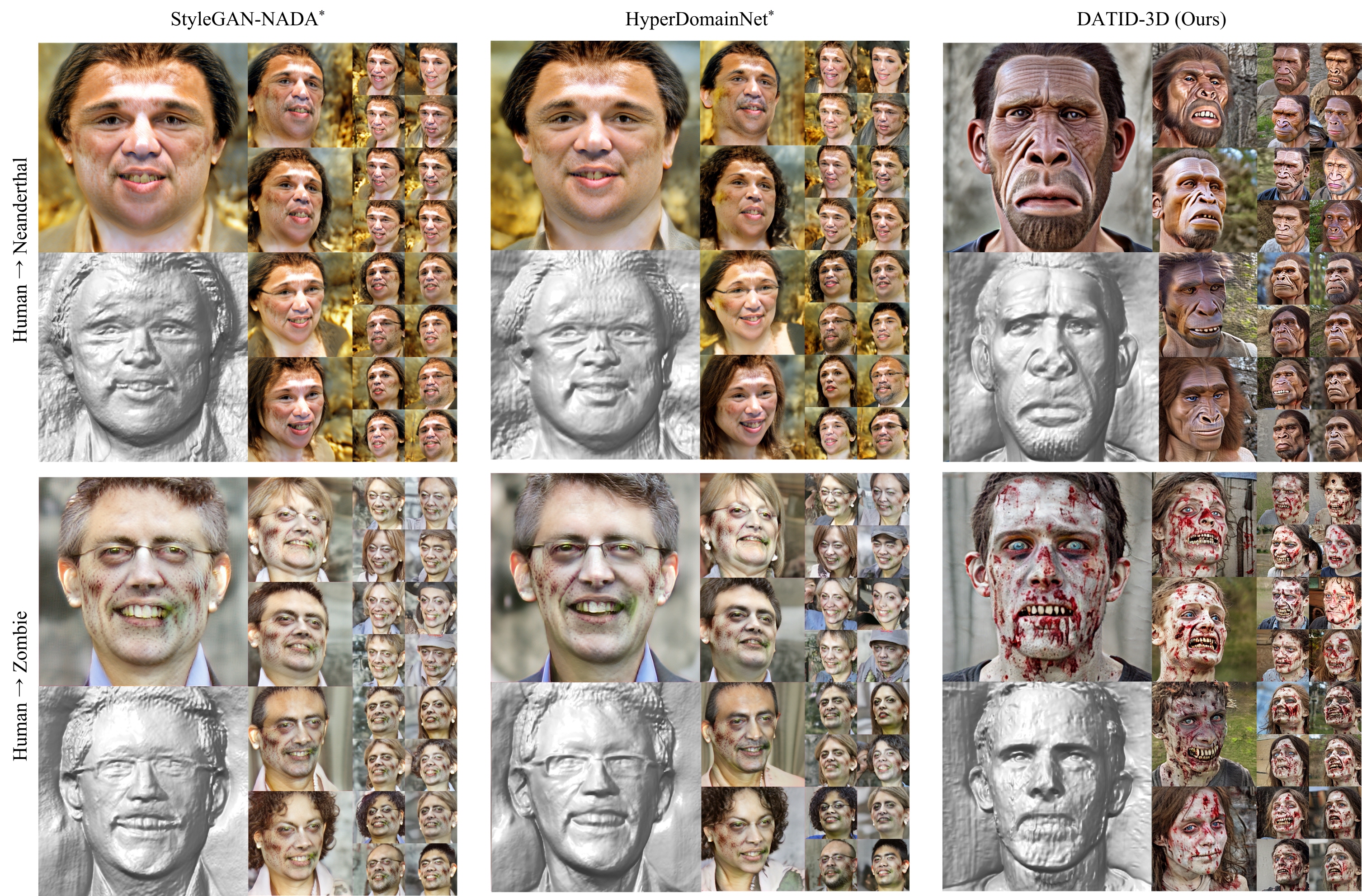}
    \caption{Qualitative comparison with the 3D extension of existing 2D text-guided domain adaptation methods (the star mark (*)).
     Our DATID-3D yielded diverse samples while other baselines did not.}
    \vspace{-1em}
    \label{fig_supp_more_comparision}
\end{figure*}

\section{Additional Results}
\label{supp_additional_results}
\subsection{Videos}

We provide an accompanying supplementary video that better visualizes and demonstrates that our methods, DATID-3D, enables the shifted generator to synthesize multi-view consistent images with high fidelity and diversity in a wide range of text-guided targeted domains at \href{https://gwang-kim.github.io/datid_3d}{gwang-kim.github.io/datid\_3d}.

\subsection{Results of text-driven 3D domain adaptation}

More results for text-driven 3D domain adaptation using the EG3D~\cite{chan2022efficient} generators pre-trained on FFHQ~\cite{karras2019style} 
or AFHQ Cats~\cite{karras2021alias, choi2020stargan} are illustrated in Figures~\ref{fig_supp_more_results1} and \ref{fig_supp_more_results2}, respectively.
Without additional images and knowledge of camera distribution, our framework allows the synthesis of diverse, high-fidelity multi-view consistent images in a wide range of text-guided domains beyond the training domains.

\subsection{Results of pose-controlled synthesis}
The results of our pose-controlled image and 3D shape synthesis in the text-guided domain are shown in Figure~\ref{fig_supp_pose_controlled}. For more results, see the provided supplementary video.

\subsection{Additional qualitative comparison results.}
In Figure~\ref{fig_supp_more_comparision}, we provide the more qualitative comparison of our method with two baselines, StyleGAN-NADA$^*$~\cite{radford2021learning} and HyperDomainNet$^*$~\cite{alanov2022hyperdomainnet}. 
By exploiting text-to-image diffusion models and adversarial training, our framework helps the shifted generator to synthesize more photorealistic and varied images.

\subsection{Additional quantitative comparison results.}
We additionally evaluate Kernel Inception Distance (KID)~\cite{binkowski2018demystifying} to calculate the distance between the distributions of generated samples and test images in the target domain because when the dataset is small, Frechet inception distance (FID)~\cite{heusel2017gans} can be easily biased while KID adopts unbiased design.
As used in the user study, EG3D pre-trained on $512^2$ images in FFHQ~\cite{karras2019style} and four text prompts converting a human face to `Pixar', `Neanderthal', `Elf' and `Zombie' styles, respectively, are employed for evaluation.
We generate 3,000 images generated through text-to-image diffusion models with a different random seed per text prompt.
As presented in Table~\ref{tab1}, our results demonstrate the superior KID as compared to the baselines. 

\begin{table}[!htb]
\caption{Quantitative comparisons with the baselines in diversity, text-image correspondence and photo realism.}\label{tab1}
\centering
\begin{adjustbox}{width=0.4\linewidth}
\begin{tabular}{lc}
\toprule
              & KID↓     \\
 \midrule
StyleGAN-NADA$^{*}$  & 0.156     \\
HyperDomainNet$^{*}$  & 0.133    \\
\textbf{Ours}  & \textbf{0.012}  \\
\bottomrule
\end{tabular}
\end{adjustbox}
\end{table}

\section{Details on Methods}
\label{supp_method_details}
\subsection{Algorithms}
\paragraph{Text-guided target dataset generation.}
The algorithm for text-guided target dataset generation is described in Algorithm~\ref{algo_target_dataset_generation}.
With each random latent vector $\zb_i \in \mathcal{Z}$ and camera parameter $\cb_i \in \mathcal{C}$,
we synthesize a source image $\xb^{\text{src}}_i$ using pre-trained 3D generator $G_\theta$. 
Then, guided by a text prompt $\y$, we perform text-guided image-to-image manipulation (\texttt{T\_I2I}) to generate $\xb^{\text{trg}}_i$ from $\xb^{\text{src}}_i$ using the text-to-image diffusion model $\epsilonb_{\phi}$.
In \texttt{T\_I2I}, we first embed 
$\xb^{\text{src}}$ into $\qb_0$ through $E^{V}$ and perturb it 
to generate $\qb^{\text{trg}}_{t_r}$ through the stochastic forward DDPM (Denoising Diffusion Probabilistic Models) process~\cite{ho2020denoising} while the return step $t_r < T_p$, where $T_p$ is the pose-consistency step.
Then, we execute the sampling process to obtain $\qb_{0}^{trg}$ from the noisy latent $\qb_{t_r}^{\text{trg}}$ using $\epsilonb_{\phi}$. $s$ controls the scale of gradients from a target prompt $y$ and a negative prompt $y_{\text{neg}}$.
Finally, the target image $\xb^{\text{trg}}$ is obtained using 
the VQGAN decoder $D^{V}$.
By repeating the above process $N$ times, we can construct a target dataset $\mathcal{D}$. 

\begin{algorithm}[h!]
    \caption{Text-guided target dataset generation\label{algo_target_dataset_generation}}
    \DontPrintSemicolon
    \SetAlgoNoLine
    \SetAlgoVlined
    \SetKwProg{Fn}{Require}{:}{}
    \SetKwProg{Fn}{Function}{:}{}
    \SetKwFunction{TextGuidedItoI}{T\_I2I}
    \KwIn{$G_\theta$, $\epsilonb_{\phi}$, $E^{V}$, $ D^{V}$, $y$, $y^{\text{neg}}$, $t_r$, $s$, $N$, *}
    \KwOut{$\mathcal{D}= \{( \cb_i, \xb_i^{\textnormal{src}}, \xb_i^{\textnormal{trg}})\}_{i=1}^{N}$}
    \Fn{\TextGuidedItoI{$\xb^{\textnormal{src}}$, $y$, $y^{\textnormal{neg}}$, $\epsilonb_{\phi}$, *}}
    {
    
    $\qb_{0}=E^{V}(\xb^{\textnormal{src}})$,  $\nb \sim \mathcal{N} (\mathbf{0,I})$ \;
    $\qb^{\textnormal{trg}}_{t_r} = \sqrt{\bar\alpha_{t_r}}\qb_{0}  + \sqrt{1 - \bar\alpha_{t_r}}\nb$ \;
    \For{$\ t = t_r, t_r-1,\ldots, 1$}{
        \small
        $\epsilonb^{\textnormal{comb}}_\phi = s\epsilonb_\phi(\qb^{\textnormal{trg}}_{t},{t}, y) + (1-s)\epsilonb_\phi(\qb^{\textnormal{trg}}_{t},{t}, y^{\textnormal{neg}})$\;
        \normalsize
        $\qb^{\textnormal{trg}}_{t-1} = \textnormal{Sampling}(\qb^{\textnormal{trg}}_{t}, \epsilonb^{\textnormal{comb}}_\phi, t)$ \;
        }
    $\xb^{\textnormal{trg}}=D^{V}(\qb^{\textnormal{trg}}_{0})$ \;
    \KwRet{$\xb^{\textnormal{trg}}$}
    }
    
    $\mathcal{D} = \{\}$ \;
    \For{$i = 1,2,\ldots, N$}{ 
        $\zb_i \in \mathcal{Z}$, $\cb_i  \in \mathcal{C}$ \;
        $\xb_i^{\textnormal{src}} = G_{\theta}(\zb_i, \cb_i)$ \;
        $\xb_i^{\textnormal{trg}}$=\TextGuidedItoI{$\xb_i^{\text{src}}$, $y$, $y^{\textnormal{neg}}$, $\epsilonb_{\phi}$, *} \;
        
        Append 
        $( \cb_i, \xb_i^{\textnormal{src}}, \xb_i^{\textnormal{trg}})$ to $\mathcal{D}$. 
    }
\end{algorithm}

\paragraph{CLIP and pose reconstruction-based filtering.}
The algorithm for CLIP and pose reconstruction-based filtering process is presented in Algorithm~\ref{algo_filtering}.
For all $(\xb_i^{\textnormal{src}}, \xb_i^{\textnormal{trg}})$ in the raw target dataset $\mathcal{D}$, we first compute the CLIP distance score ${d}_{\text{CLIP}}$ between the target image $\xb^{\text{trg}}_i$ and the target prompt $y$.
If ${d}_{\text{CLIP}} > \alpha$, then replace $\xb^{\text{trg}}_i$ with a new one through \texttt{T\_I2I} and repeat the CLIP-based filtering again. 
Otherwise, we convert 
$\xb^{\text{trg}}_i$ to a reconstructed image $\xb^{\text{rec}}_i$ using the Reconstructor latent diffusion $\epsilonb_{\phi^{\textnormal{rec}}}$. 
Then, we calculate the pose difference score ${d}_{\textnormal{pose}}$ between the reconstructed image $\xb^{\text{rec}}_i$ and the target image $\xb^{\text{trg}}_i$.
If ${d}_{\textnormal{pose}}> \beta$, then replace $\xb^{\text{trg}}_i$ with a new one through \texttt{T\_I2I} and repeat the CLIP-based filtering again. 
Otherwise, we can finish the filtering for $\xb_i^{\textnormal{trg}}$ and save a set of $( \cb_i, \xb_i^{\textnormal{src}}, \xb_i^{\textnormal{trg}})$ to $\mathcal{D}_f$. 
In practice, it sometimes takes a too long time to repeat the process until 
$\xb_i^{\textnormal{trg}}$ passes, we only repeat it by $K_{\text{f}}$ times, which was set to 5 for our experiments.

 \begin{algorithm}[h!]
    \caption{CLIP and pose reconstruction-based filtering\label{algo_filtering}}
    \DontPrintSemicolon
    \SetAlgoNoLine
    \SetAlgoVlined
    \SetKw{Continue}{continue}
    \SetKw{Break}{break}
    \SetKwProg{Fn}{Require}{:}{}
    \KwIn{$\mathcal{D}$, $\epsilonb_{\phi^{\textnormal{rec}}}$, $\epsilonb_{\phi}$, $y^{\textnormal{src}}$ , $y$ $y^{\text{neg}}$, $N$, *}
    \KwOut{$\mathcal{D}_f= \{(\cb_i, \xb_i^{\textnormal{src}}, \xb_i^{\textnormal{trg}})\}_{i=1}^{N}$}
    $\mathcal{D}_f = \{\}$ \;
    \For{$i = 1,2,\ldots, N$}{ 
        $(\xb_i^{\textnormal{src}}, \xb_i^{\textnormal{trg}}) \in \mathcal{D}$ \;
        
        \While{True}{
            \uIf{$ {d}_{\textnormal{CLIP}}(\xb_i^{\textnormal{trg}}, {y}) > \alpha$}{
                $\xb_i^{\textnormal{trg}}$=\TextGuidedItoI{$\xb_i^{\textnormal{src}}$, $y$, $y^{\textnormal{neg}}$, $\epsilonb_{\phi}$, *} \;
                \Continue \;
              }
            \Else{
            \small
            $\xb_i^{\textnormal{rec}} =$\TextGuidedItoI{$\xb_i^{\textnormal{src}}$, $y^{\textnormal{src}}$, $\textnormal{None}$, $\epsilonb_{\phi^{\textnormal{rec}}}$, *}  \;
            \normalsize    \uIf{${d}_{\textnormal{pose}}(\xb_i^{\textnormal{rec}}, \xb_i^{\textnormal{src}}) > \beta$}{
                $\xb_i^{\textnormal{trg}}$=\TextGuidedItoI{$\xb_i^{\textnormal{src}}$, $y$, $y^{\textnormal{neg}}$, $\epsilonb_{\phi}$, *} \;
                \Continue \;
              }
                \Else{
                    \Break \;
                }
            }
        }
        Append 
        $( \cb_i, \xb_i^{\textnormal{src}}, \xb_i^{\textnormal{trg}})$ to $\mathcal{D}_f$.

    }

\end{algorithm}

\paragraph{Diversity-preserved domain adaptation.}

The algorithm for diversity-preserved domain adaptation is provided in Algorithm~\ref{algo_domain_adaptation}.
We first clone the pre-trained 3D generator $G_\theta$ to $G_{\theta'}$ and initialize pose-conditioned discriminator $D_\psi$. 
For $i=1,2,..., N$, we first sample a random latent vector and camera parameter. 
Then, we compute ADA loss for the generator $\mathcal{L}^{\theta'}_{\text{ADA}}$ with generated images $G_{\theta'}(\zb_i, \cb_i)$ using $D_\psi$ and the stochastic non-leaking augmentation $A$.
Also, we calculate the density regularization loss $\mathcal{L}_{\text{den}}$ with randomly chosen points $v$ from the volume $\mathcal{V}$ for each rendered scene.
With these two losses, the generator is updated.
Next, we compute ADA losses for the discriminator, $\mathcal{L}^{\psi,\textnormal{fake}}_{\text{ADA}}$ and  $\mathcal{L}^{\psi,\textnormal{real}}_{\text{ADA}}$, with generated images $G_{\theta'}(\zb_i, \cb_i)$ and real targets $\xb_i^{\textnormal{trg}}$, respectively.
Combining these two losses, the discriminator is updated.
We repeat this process for $K$ epochs.

\begin{algorithm}[h!]
    \caption{Diversity-preserved domain adaptation\label{algo_domain_adaptation}}
    \DontPrintSemicolon
    \SetAlgoNoLine
    \SetAlgoVlined
    \SetKwProg{Fn}{Require}{:}{}
    \SetKwProg{Fn}{Function}{:}{}
    \SetKwFunction{TextGuidedItoI}{T\_I2I}
    \KwIn{$G_\theta$ (pre-trained 3D generator),  $\mathcal{D}_f$ (filtered dataset), $N$ (Number of data), $K$ (total number of epochs),  $A$ (stochastic non-leaking augmentation), $f$, *}
    \KwOut{$G_{\theta'}$}
    $G_{\theta'} \leftarrow \textnormal{clone}(G_{\theta})$, 
    $D_\psi \leftarrow \textnormal{Initialize\_}D$ \;
    \For{$k = 1,2,\ldots, K$}{ 
        \For{$i = 1,2,\ldots, N$}{ 
            $\zb_i \in \mathcal{Z}$, $\cb_i  \in \mathcal{C}$, $v_i \in \mathcal{V}$ \;
            
            \tcp{Step 1: Update $G_{\theta'}$}
            $\mathcal{L}^{\theta'}_{\text{ADA}}=-f(D_\psi(A(G_{\theta'}(\zb_i, \cb_i)), \cb_i)$ \;
            
            $\mathcal{L}^{\theta'}_{\text{den}}=\|\sigma_{\theta'}(v_i) - \sigma_{\theta'}(v_i+\delta v_i)\|$ \;
            
            $\theta' \leftarrow \textnormal{Update\_}G(\theta',\mathcal{L}^{\theta'}_{\text{ADA}} + \lambda_{\textnormal{den}}\mathcal{L}_{\text{den}})$
            
            \tcp{Step 1: Update $D_{\psi}$}
            $\mathcal{L}^{\psi,\textnormal{fake}}_{\text{ADA}}=f(D_\psi(A(G_{\theta'}(\zb_i, \cb_i)), \cb_i)$
            
            $(\cb_i, \xb_i^{\textnormal{trg}}) \in \mathcal{D}$ \;
            $\mathcal{L}^{\psi,\textnormal{real}}_{\text{ADA}}= f(-D_{\psi}(A(\xb^{\text{trg}}), \cb_i) $\\
            \quad\quad\quad\quad\quad$+\lambda\|\nabla D_{\psi}(A(\xb^{\text{trg}}), \cb_i)\|^2)$
            
            $\psi \leftarrow
            \textnormal{Update\_}D(\psi,\mathcal{L}^{\psi,\textnormal{fake}}_{\text{ADA}} + \mathcal{L}^{\psi,\textnormal{real}}_{\text{ADA}})$

        }
    }

\end{algorithm}

\section{Implementation Details}
\label{supp_implementation_details}
\subsection{3D generative model}
We adopt EG3D~\cite{chan2022efficient}, the state-of-the-art 3D generative model pre-trained on $512^2$ images in FFHQ~\cite{karras2019style} and AFHQ Cats~\cite{karras2021alias, choi2020stargan} as our source generator.
Its generator is composed of a backbone, decoder, volume rendering, and super-resolution parts.
The backbone consists of the StyleGAN2 generator~\cite{karras2020analyzing} and a mapping network with 8 hidden layers.
The decoder is constructed as an MLP with a single hidden layer with soft plus activation and neural rendering~\cite{mildenhall2020nerf} of features~\cite{niemeyer2021giraffe} using two-pass importance is utilized.
The super-resolution module is implemented with two StyleGAN2 blocks with modulated convolutions.
EG3D's discriminator is based on a StyleGAN2 discriminator with two changes, dual discrimination, and camera pose-conditioning.

\subsection{Text-to-image diffusion model}
We employ Stable diffusion~\cite{rombach2022high} as our text-to-image diffusion model.
It is a latent-based diffusion model and leverages a pre-trained 123M CLIP ViT-L/14~\cite{radford2021learning} text encoder to provide the model with the condition of text prompts. 
The diffusion model where 860M UNet~\cite{ronneberger2015u} with the text encoder are combined is lightweight and enables text-to-image synthesis on GPU at 10GB VRAM.
We use Stable diffusion v1.4, where 977k steps were taken at 512$\times$512 images paired with text captions from a subset of the LAION-5B~\cite{schuhmann2022laion} database. 

For the diffusion sampling method, we choose PLMS~\cite{liu2022pseudo}, one of the state-of-the-art sampling methods, accelerating the diffusion process with high quality.
We set the number of inference steps to 50, which enables us to generate a high-quality image in 1$\sim$2 seconds. 
We generally set $y^{\text{neg}}$ to None. 
Also, we generally set the return step $t_r$ and the guidance scale $s$ to 700 and 10, respectively.

\subsection{Pose-extractor}
As a pose-extractor, we use 6DRepNet~\cite{hempel20226d} that demonstrates the state-of-the-art performance on BIWI~\cite{borghi2017poseidon} head pose estimation benchmark.
This model predicts a pose vector on images that includes yaw, pitch, and roll vectors.
We found that this model works well on both FFHQ~\cite{karras2019style} and AFHQ Cats~\cite{karras2021alias, choi2020stargan} images, thus we use the model for both types of images. 

\subsection{Fine-tuning details}

We fine-tune the 3D generative models with a batch size of 20 until the models see 50,000$\sim$200,000 images. 
We use a learning rate of 0.002 for both the generator and discriminator.
For the discriminator's input, we blur images, progressively diminishing the blur degree following~\cite{karras2021alias, chan2022efficient} and don't use style mixing during training.
We use ADA loss combined with R1 regularization with $\lambda=5$. 
We set the strength of density regularization $\lambda_{\text{den}}$ to $0.25$.

\subsection{3D shape visualization}
To visualize 3D shapes, we first extract iso-surfaces from the density field using marching cubes following \cite{chan2022efficient}. 
Then, we view the 3D surfaces using UCSF Chimerax~\cite{chimerax}.



\begin{table*}[!htb]
\centering
\caption{List of full text prompts corresponding to each text prompt.}\label{tab_text_prompt}%
\begin{adjustbox}{width=0.9\linewidth}
\begin{tabular}{ccc}
\toprule
\multicolumn{1}{l}{Source data type} & Concise prompt & Full text prompt \\
\midrule
\multirow{17}{*}{FFHQ} & Lego & \textit{"a 3D render of a head of a lego man 3D model"} \\
 & Greek Statue & \textit{"a FHD photo of a white Greek statue"} \\
 & Pixar & \textit{"a 3D render of a face in Pixar style"} \\
 & Orc & \textit{"a FHD photo of a face of an orc in fantasy movie"} \\
 & Elf & \textit{"a FHD photo of a face of a beautiful elf with silver hair in live action movie"} \\
 & Neanderthal & \textit{"a FHD photo of a face of a neanderthal"} \\
 & Skeleton & \textit{"a FHD photo of a face of a skeleton in fantasy movie"} \\
 & Zombie & \textit{"a FHD photo of a face of a zombie"} \\
 & Masquerade & \textit{"a FHD photo of a face of a person in masquerade"} \\
 & Peking opera & \textit{"a FHD photo of face of character in Peking opera with heavy make-up"} \\
 & Tekken & \textit{"a 3D render of a Tekken game character"} \\
 & Ston golem & \textit{"a 3D render of a stone golem head in fantasy movie"} \\
 & Devil & \textit{"a FHD photo of a face of a devil in fantasy movie"} \\
 & Baby & \textit{"a FHD photo of a face of a cute baby"} \\
 & Super Mario & \textit{"a 3D render of a face of Super Mario"} \\
 & Hobbit & \textit{"a FHD photo of a face of Hobbit in Lord of the Rings "} \\
 & Yoda & \textit{"a FHD photo of a face of Yoda in Star Wars"} \\
 \midrule
\multirow{11}{*}{AFHQ Cats} & Golden statue & \textit{"a photo of a face of an anmial golden statue"} \\
 & Madagascar character & \textit{"a 3D render of a face of a animal animation character in Madagascar style"} \\
 & Eevee in Pokemon & \textit{"a 3D render of a face of an eevee in Pokemon"} \\
 & Lion in Zootopia style & \textit{"a 3D render of a face of a lion in Zootopia style"} \\
 & Cat in Zootopia style & \textit{"a 3D render of a face of a cat in Zootopia style"} \\
 & Wolf in Zootopia style & \textit{"a 3D render of a face of a wolf in Zootopia style"} \\
 & Fox in Zootopia style & \textit{"a 3D render of a face of a fox in Zootopia style"} \\
 & Sheep in Zootopia style & \textit{"a 3D render of a face of a sheep in Zootopia style"} \\
 & Pig in Zootopia style & \textit{"a 3D render of a face of a pig in Zootopia style"} \\
 & Hamster in Zootopia style & \textit{"a 3D render of a face of a hamster in Zootopia style"} \\
 & Racoon in Zootopia style & \textit{"a 3D render of a face of a racoon in Zootopia style"} \\
\bottomrule
\end{tabular}
\end{adjustbox}
\end{table*}

\subsection{Text prompts}
In the main paper and supplementary, we use a concise text prompt to refer to each text prompt.
Full-text prompts corresponding to each concise text prompt are summarized in Table~\ref{tab_text_prompt}.

\section{Experimental Details}
\label{supp_experimental_details}
\subsection{Evaluation details}
\paragraph{Baselines.}

In StyleGAN-NADA$^*$ that is a 3D extended version of StyleGAN-NADA~\cite{radford2021learning}, we fine-tune the 3D generator $G^{\theta}$ with the directional CLIP loss as follows:
\small
\begin{align}
    \mathcal{L}^{\theta}_{\text{direction}} = 1 - \frac{\langle \Delta I, \Delta T\rangle}{\|\Delta I\|\|\Delta T\|},
\end{align}
\normalsize
where \( \Delta I =  E_I^{{C}}(\xb^{\text{gen}}) - E_I^{{C}}(\xb^{\text{src}}), \Delta T =  E_T^{{C}}{(y^{\text{tar}}}) - E_T^{{C}}({y^{\text{src}}}) \).
We implement the loss and optimization part based on the official StyleGAN-NADA codebase~\cite{radford2021learning}.

In HyperDomainNet$^*$ that is a 3D extended version of HyperDomainNet~\cite{alanov2022hyperdomainnet}, in-domain angle consistency loss $\mathcal{L}_{\text{indomain}}$ is added to the directional CLIP loss for preserving the CLIP similarities among images before and after domain adaptation.
\small
\begin{align}
\mathcal{L}^{\theta}_{\text{indomain}} = \sum_{i, j}^n(\langle E^C_I(\xb^{\text{gen}}_i), E^C_I(\xb^{\text{gen}}_j)\rangle-\langle E^C_I(\xb^{\text{src}}_i), E^C_I(\xb^{\text{src}}_j)\rangle)^2,
\end{align}
\normalsize
We implement the loss part based on the official HyperDomainNet~\cite{alanov2022hyperdomainnet}.

\paragraph{KID.}
Based on the StyleGAN3~\cite{karras2021alias} codebase implementation, we calculate Kernel Inception Distance (KID) between 50,000 produced images and 3,000 training images.

\paragraph{User study.}
For the user study, we collect 9,000 votes from 75 people using a survey platform.
We adapt the generator using each method for four text prompts converting a human face to `Pixar', `Neanderthal', `Elf' and `Zombie' styles as these prompts are used in the previous work, StyleGAN-NADA~\cite{radford2021learning}. 
Then, for each text prompt, we sample 30 images from each generator and put the results of each method side-by-side. To quantify opinions, we requested users to rate the perceptual quality on a scale of 1 to 5 for 3 questions as we introduced in the main text.
Finally, we report the mean of each score for each method, respectively.

\paragraph{Non-adversarial fine-tuning.}
One generator per instance is optimized like StyleGAN-NADA*~\cite{radford2021learning}, but the difference is that the guidance is from CLIP image encoding of the target images that were generated from text-to-image diffusion models, not CLIP text encoding.

\subsection{3D GAN inversion}
For single-view manipulated 3D reconstruction, we invert a real image into the latent vector $w$ in $\mathcal{W}^+$ space. 
To achieve this, we obtain the camera parameter $c$ with pre-trained pose extractor~\cite{deng2019arcface, hempel20226d} and we initialize $w$ as a mean of 10,000 $w$s that are mapped from $z$s which are randomly sampled from Normal distribution.
Then, we generate a images with 3D generator and compute a feature distance between the generated image and the real image using VGG-19 network.
Then, using Adam optimizer~\cite{kingma2014adam}, we update the $w$ minimizing the feature distance for 1,000 steps.

\begin{figure}[!t]
\captionof{table}{High diversity is ensured by sampling more target images (large $n$) with our CLIP and pose reconstruction-based filtering.}\label{tab4_number_of_data}
    \vspace{-1em}
\begin{minipage}{.33\linewidth}
 \centering 
\begin{adjustbox}{width=0.90\linewidth}
\begin{tabular}{lc}
\toprule
 & KID ↓ \\
 \midrule
$n=100$ & 0.024 \\
$n=500$ & 0.015 \\
$n=1000$ & 0.013  \\
$n=3000$  & {0.012}  \\
\bottomrule
\end{tabular}
\end{adjustbox}
\end{minipage}\hfill
\begin{minipage}{.66\linewidth}
\centering
\includegraphics[width=\linewidth]{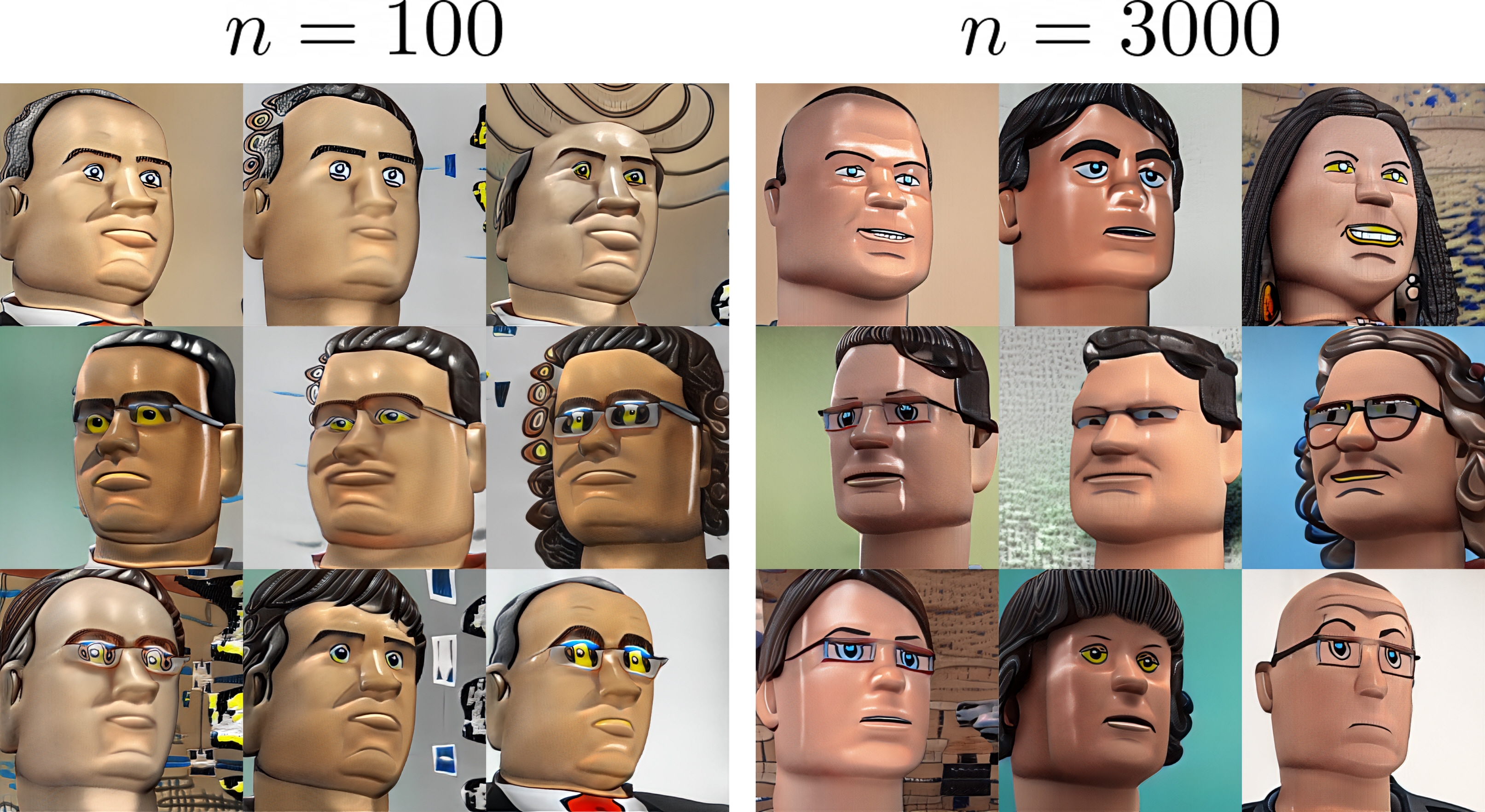}
\end{minipage}
\end{figure}

\begin{figure}[!t]
\captionof{table}{Trade-off between image-text correspondence $d_{\text{CLIP}}$ and pose-consistency $d_{\text{pose}}$ related to the return step $t_r$.}\label{tab_supp_return_step}
    \vspace{-1em}
\begin{minipage}{.35\linewidth}
 \centering 
\begin{adjustbox}{width=0.90\linewidth}
\begin{tabular}{lcc}
\toprule
$t_r$ & $d_{\text{pose}}$↓ & $d_{\text{CLIP}}$↓ \\
\midrule
500 & \textbf{8.133} & 0.689 \\
600 & 23.381 & 0.672 \\
700 & 86.516 & 0.657 \\
800 & 263.081 & 0.654 \\
900 & 327.478 & \textbf{0.652} \\
\bottomrule
\end{tabular}
\end{adjustbox}
\end{minipage}\hfill
\begin{minipage}{.65\linewidth}
\centering
\includegraphics[width=\linewidth]{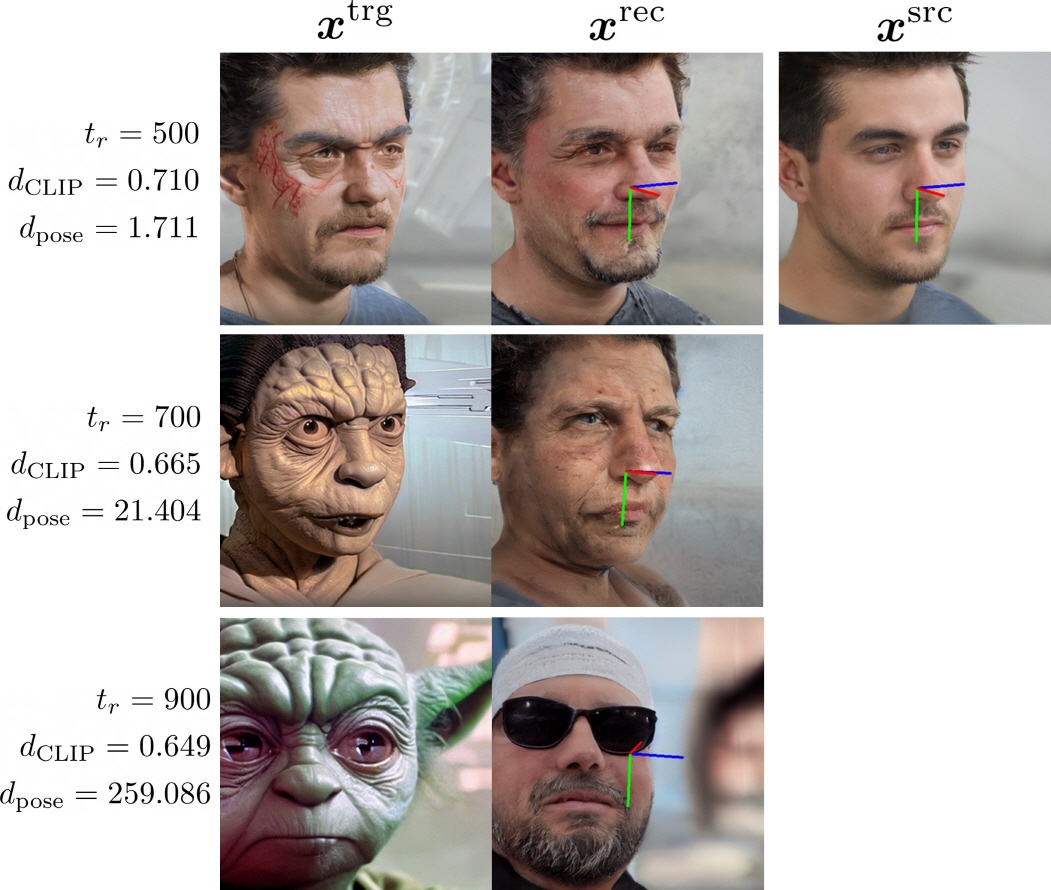}
\end{minipage}
\end{figure}

\begin{figure}[!t]
    \centering
    \vspace{-0.5em}
    \includegraphics[width=0.9\linewidth]{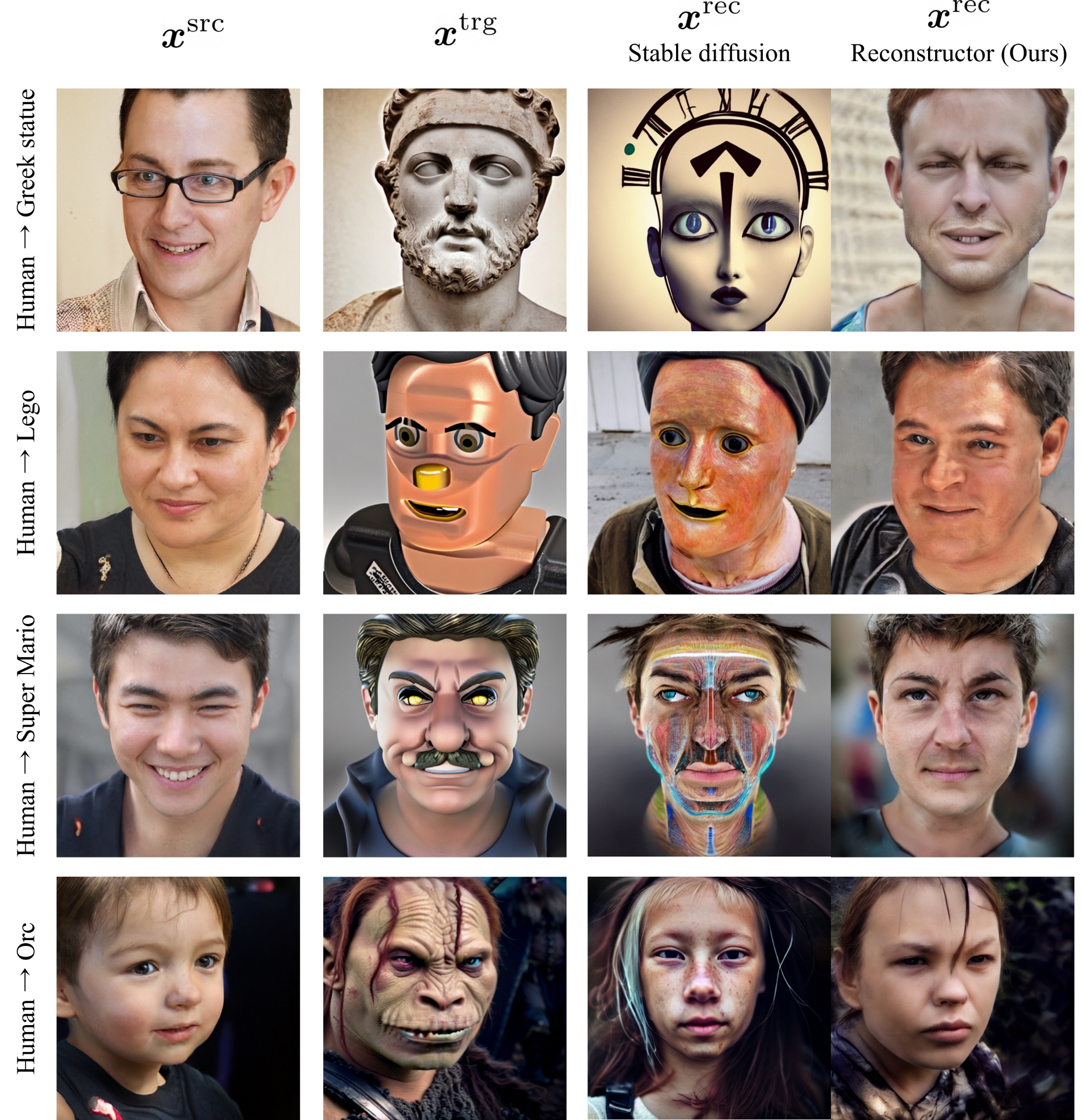}
    \caption{Reconstructor successfully converted the target images into the images in the source domain (Human face) without unrealistic artifacts.}
    \label{fig_supp_reconstructor}
\end{figure}

\begin{figure}[!t]
    \centering
    \includegraphics[width=0.85\linewidth]{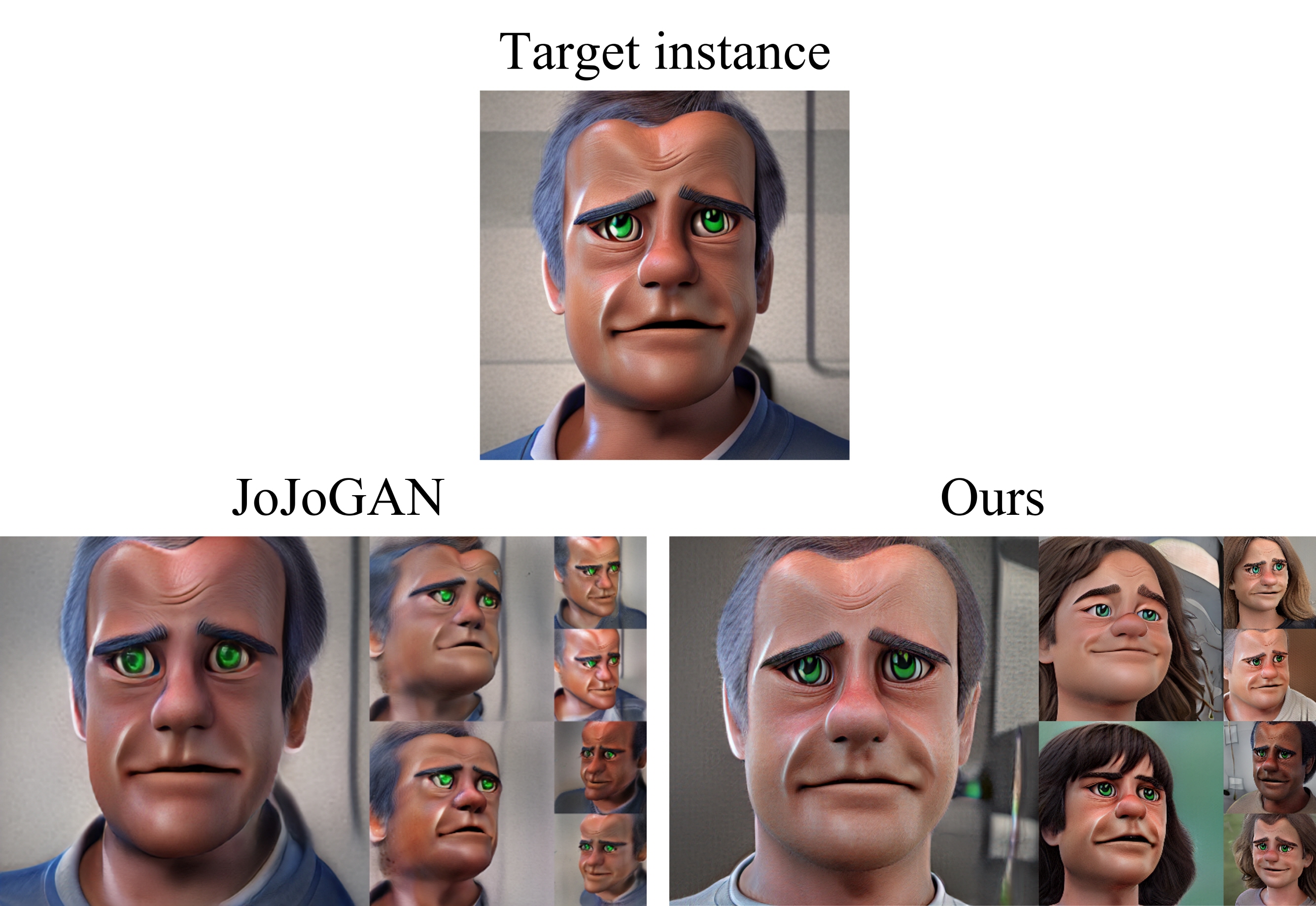}
    \vspace{-1em}
    \caption{Comparison of our one-shot fine-tuning method with JoJoGAN~\cite{chong2022jojogan}, the state-of-the-art one-shot stylization method. Our method shows more diverse images with higher quality.}
    \vspace{-1em}
    \label{fig_supp_one_shot}
\end{figure}

\section{Additional Ablation Studies}
\label{supp_additional_ablation}
\paragraph{Number of samples.}
We also analyzed the diversity, image quality, and training time depending on the number of samples.
According to the quantitative (table) and qualitative (figures) results in Table~\ref{tab4_number_of_data}, more sampled target images lead to improved image quality and diversity.

\paragraph{Trade-off related to return step.}
The return step $t_r$ is one of the important hyperparameters that determines the degree of text changes guided by image-to-image manipulation.
We identified that there is a trade-off between image-text correspondence and pose consistency related to the return step.
According to the quantitative (table) and qualitative (figures,  `Human face' $\rightarrow$ `Yoda') results in Table~\ref{tab_supp_return_step}, higher return step results in a lower CLIP distance score, but a higher pose difference score.
Thus, we set $t_r$ to 600$\sim$700 depending on the text prompts.

\paragraph{Effectiveness of the Reconstructor.}
Here, we compare the reconstruction performance between our proposed Reconstructor and original Stable diffusion.
As a text prompt, the Stable diffusion uses ``A photo of a human face'' while the Reconstructor use ``A photo of a <s> human face'' that includes a specifier word.
Our goal is to translate the manipulated target image back to the image in the source domain.
As shown in Figure~\ref{fig_supp_reconstructor}, the results from the stable diffusion reveal loss of pose information or artificial distortions because of its highly stochastic nature, whereas the Reconstructor successfully transforms the target images into the images in the source domain (Human face).

\paragraph{Effectiveness of one-shot fine-tuning using text-to-diffusion model.}
Here, we compare our one-shot fine-tuning method with the 3D extension of the state-of-the-art method of one-shot stylization for 2D generative models, JoJoGAN~\cite{chong2022jojogan}. 
We add the camera sampling procedure to the domain adaptation pipeline in JoJoGAN.
As presented in Figure~\ref{fig_supp_one_shot}, our one-shot fine-tuning method shows superior image quality and diversity for 3D generative models while the results from JoJoGAN severely overfit the target images.

\begin{figure}[!tb]
    \centering
    \includegraphics[width=0.85\linewidth]{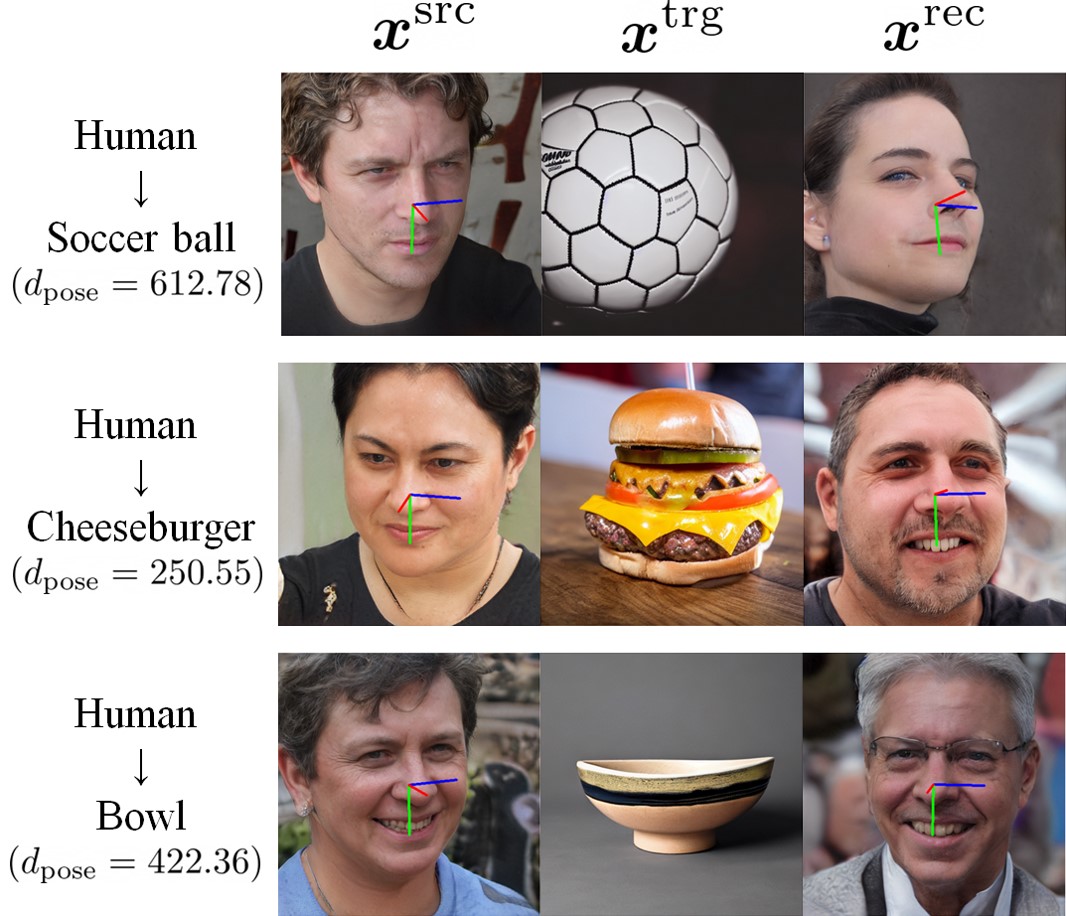}
    \vspace{-1em}
    \caption{Manipulation to rotation-invariant objects shows high pose-difference scores. }
    \vspace{-1em}
    \label{fig_supp_limitation}
\end{figure}

\section{Discussion}
\label{supp_additional_discussion}
\paragraph{Limitation.}
We discovered that maintaining posture information in the target images created in Stage 1 is a crucial requirement for a successful text-driven 3D domain adaption.
There are, however, certain inevitable circumstances that fit this requirement.
The target object being rotation-invariant or in 2D space is one of the situations when pose information is lost.
As shown in Figure~\ref{fig_supp_limitation}, image manipulation of `Human face' $\rightarrow$ `Cheeseburger', `Human face' $\rightarrow$ `Soccer ball' and `Human face' $\rightarrow$ `Bowl' reports high pose-difference score, failing domain adaptation with flattened 3D shapes as described in Figure 11 in the main text. 

Also, the supervision of our text-guided domain adaptation depends on the power of text-to-image diffusion models.
So, the limitation of the chosen diffusion models is inherited in our pipeline.
In this work, we adopt Stable diffusion~\cite{rombach2022high}.
According to the Stable diffusion model card, a limitation of the model includes falling short of achieving (1) complete photorealism, (2) compositionality, (3) proper face generation, (4) generating images with other languages except for English, and so on.
These limitations can affect our performance of ours.

\paragraph{Diversity.}
The diversity of generated samples from the shifted generator depends on the diversity of the target dataset. 
For example, the target images from the text prompt `Human face' $\rightarrow$ `Super Mario' will be less diverse and more biased to the specific concept than the target images from `Human face' $\rightarrow$ `Pixar'.
Thus, the domain adaptation results using the text prompt `Human face' $\rightarrow$ `Super Mario' are also less diverse than the results using `Human face' $\rightarrow$ `Pixar'.
Also, as analyzed in \cite{karras2020training}, transfer learning of the generative models succeeds only when the target dataset has comparable or less diverse than the source dataset. 

\paragraph{Social Impacts}
DATID-3D enables the generation of high-quality 3D samples in the text-guided domain as well as single-shot manipulated 3D reconstruction without artistic skills.
Nevertheless, these can be applied maliciously to produce visuals that make people feel unpleasant or aggressive. This involves creating images that people are likely to find upsetting, frightening, or insulting, as well as information that reinforces stereotypes from the past or present.
According to the Stable diffusion~\cite{rombach2022high} model card, a misuse of the model includes 
(1) creating inaccurate, hurtful, or otherwise offensive depictions of individuals, their environment, cultures, and religions, (2) intentionally spreading stereotypical portrayals or discriminatory material, (3) impersonating individuals without their consent, (4) sexual content without viewer's permission, (5) depictions of horrifying violence and gore and so on. 
We thus strongly urge people to use our approach wisely and for the proper intended goals.


\end{document}